\documentclass[journal]{IEEEtran}
\pdfoutput=1
\usepackage{mathrsfs}
\usepackage{pifont}
\usepackage{bbding}
\usepackage{tipa}
\usepackage{color}
\usepackage{cite}
\usepackage{epsfig}
\usepackage{graphicx}
\usepackage{url}
\usepackage{amsfonts}
\usepackage{multicol}
\usepackage{float}
\usepackage{times}
\usepackage{psfrag}
\usepackage{subfigure}
\usepackage{stfloats}
\usepackage{amsmath}
\usepackage{citesort}
\usepackage{footnote}
\usepackage{array}
\usepackage{stmaryrd}
\usepackage{amssymb}
\usepackage{epic}
\usepackage{graphicx}
\usepackage{curves}
\usepackage{mathbbol}
\usepackage{balance}
\usepackage{algorithm}
\usepackage{algorithmic}
\usepackage{multirow}
\usepackage{amsmath}
\usepackage{xcolor}
\usepackage{graphics}
\usepackage{epstopdf}
\usepackage{enumerate}
\usepackage{caption2}
\usepackage{booktabs}
\usepackage{bbm}
\usepackage{tcolorbox}

\newtheorem{remark}{Remark}

\begin{document}

\title{Decentralized Federated Reinforcement Learning for User-Centric Dynamic TFDD Control}
%
%
\author{Ziyan Yin,~\IEEEmembership{Student Member,~IEEE,} Zhe Wang,~\IEEEmembership{Member,~IEEE,} Jun Li,~\IEEEmembership{Senior Member,~IEEE,} Ming Ding,~\IEEEmembership{Senior Member,~IEEE,} Wen Chen,~\IEEEmembership{Senior Member,~IEEE,} and Shi Jin,~\IEEEmembership{Senior Member,~IEEE}
%

%
 \thanks{Ziyan Yin and Jun Li are with the School of Electronic and Optical Engineering, Nanjing University of Science and Technology, Nanjing 210094, China (e-mail: \{ziyan.yin, jun.li\}@njust.edu.cn).}
 \thanks{Zhe Wang is with the School of Computer Science and Engineering, Nanjing University of Science and Technology, Nanjing 210094, China (e-mail: zwang@njust.edu.cn).}
 \thanks{Ming Ding is with Data61, CSIRO, Sydney, Australia (e-mail: ming.ding@data61.csiro.au).}
  \thanks{Wen Chen is with the Department of Electronic Engineering, Shanghai Jiao Tong University, Shanghai 200240, China (e-mail: wenchen@sjtu.edu.cn).}
  \thanks{Shi Jin is with the National Mobile Communications Research Laboratory, Southeast University, Nanjing 210096, China (e-mail: jinshi@seu.edu.cn).}
}
\maketitle

\begin{abstract}
The explosive growth of dynamic and heterogeneous data traffic brings great challenges for 5G and beyond mobile networks. To enhance the network capacity and reliability, we propose a learning-based dynamic time-frequency division duplexing (D-TFDD) scheme that adaptively allocates the uplink and downlink time-frequency resources of base stations (BSs) to meet the asymmetric and heterogeneous traffic demands while alleviating the inter-cell interference. We formulate the problem as a decentralized partially observable Markov decision process (Dec-POMDP) that maximizes the long-term expected sum rate under the users' packet dropping ratio constraints. In order to jointly optimize the global resources in a decentralized manner, we propose a federated reinforcement learning (RL) algorithm named federated Wolpertinger deep deterministic policy gradient (FWDDPG) algorithm. The BSs decide their local time-frequency configurations through RL algorithms and achieve global training via exchanging local RL models with their neighbors under a decentralized federated learning framework. Specifically, to deal with the large-scale discrete action space of each BS, we adopt a DDPG-based algorithm to generate actions in a continuous space, and then utilize Wolpertinger policy to reduce the mapping errors from continuous action space back to discrete action space. Simulation results demonstrate the superiority of our proposed algorithm to benchmark algorithms with respect to system sum rate.
\end{abstract}


\markboth{submitted to IEEE Journal of selected topics in signal processing}
{Shell \MakeLowercase{\textit{et al.}}: Bare Demo of IEEEtran.cls for IEEE Communications Society Journals}

\section{Introduction}
Driven by the burgeoning demands of various services coming from smart cities and industries, 5th generation (5G) and beyond wireless communication systems are facing the challenges of diverse quality-of-service (QoS) requirements~\cite{Resource2022Setayesh,Privacy2022Niu,ASurvey2021}. The conventional ``one-size-fit-all" network infrastructure may not be able to simultaneously meet the heterogeneous service requirements. Network slicing has been proposed to ``slice" the mobile infrastructure into multiple logical networks, which provides flexible network services in a cost-efficient way~\cite{Federated2020Yingyu}~\cite{Design2022Hsieh}. The key problem for network slicing is to dynamically and efficiently allocate the computation and communication resources, e.g., computing frequencies~\cite{Federated2020Yingyu}, transmit power \cite{Deep2021Messaoud}\cite{Robust2022Xu}, radio spectrum~\cite{Deeo2019Qi} and transmission time~\cite{On2018Shrivastava}, to meet various and even conflicting QoS demands.

Time division duplexing (TDD), as a typical application of network slicing, is able to accommodate asymmetric traffic demands in the uplink (UL) and downlink (DL) by allowing the UL and DL traffic to operate in different subframes \cite{Requirements20173GPP}. The TDD system can be mainly classified into two categories: static TDD (S-TDD) and dynamic TDD (D-TDD). For S-TDD \cite{Dynamic2012Shen,Dynamic2020Kim,Joint2021Song}, all base stations (BSs) adopt the same and synchronized UL and DL subframe configurations, which however may not be efficient if the traffic demands are dynamic and asymmetric across the cells. To improve the resource utilization efficiency, D-TDD is proposed, where BSs can adopt different subframe configurations.
However, D-TDD suffers from additional inter-cell interference due to the asynchronous transmissions, i.e., the UL/DL transmissions in a cell may interfere with the DL/UL transmissions in its neighboring cells~\cite{Centralized2021Ghermezcheshmeh}.
To alleviate the inter-cell interference, the BSs can be divided into different clusters \cite{Analysis2018J.}, where the BSs within each cluster adopt the same subframe configuration.
Another interference alleviation approach is to adjust the wireless signal transmission strategies, i.e.,  interference cancellation~\cite{Dynamic2020Kim}~\cite{On2016M.}, power control~\cite{Joint2021Lee}\cite{Game2021Sheemar}   and beamforming~\cite{BiDirectionalBT2018Jayasinghe,3D2020Rachad,Joint2021Lee}, where the BSs cooperatively optimize their signal transmission strategies via convex optimization or heuristic algorithms. For this type of approach, the subframe configuration is usually selected from pre-defined candidates, e.g., the seven subframe configurations of 3GPP~\cite{Further20123GPP}, without adapting to the real-time traffic demands.

The network traffic demands and channel states are highly dynamic and unpredicted in D-TDD systems, making it costly to design the  adaptive subframe configurations by the conventional model-based optimization methods. Advanced model-free methods such as single-agent reinforcement learning (RL)~\cite{deep2017He}~\cite{Software2017He} and multi-agent reinforcement learning (MARL)~\cite{DeepRL2020Tang,QoE2019Tsai,Dynamic2014Wang} have been recently applied to solve the sequential resource allocation problems in complex and dynamic wireless networks, where the agents can learn the policy in a trial-and-error manner.
There are two main types of MARL approaches for designing the subframe configurations in the D-TDD system: centralized MARL~\cite{DeepRL2020Tang} and decentralized MARL \cite{QoE2019Tsai}\cite{Dynamic2014Wang}. The subframe configuration in \cite{DeepRL2020Tang} depends on the coordination of a centralized controller.
The BSs in \cite{QoE2019Tsai}\cite{Dynamic2014Wang} independently make local subframe configuration decisions, while treating other BSs as part of the environment.

Based on the aforementioned literature, there are two challenges left unsolved. The first challenge is to design the D-TDD scheme to meet the heterogeneous QoS demands of different user equipment (UE) types. In the existing literature, most of the D-TDD subframe configurations are cell-centric, where each BS allocates the number of UL/DL subframes depending on the average UL/DL data traffic inside this cell without further differentiating the resource demands of the specific UEs. However, the data traffic patterns and the QoS requirements may vary significantly for different UE types in a heterogeneous network, which has been largely overlooked in the existing literature. To satisfy the user-centric heterogeneous QoS demands, we propose a learning-based dynamic time-frequency division duplexing (D-TFDD) framework.
The second challenge is to jointly optimize the resources for local traffic adaptation and global interference alleviation without collecting the private states from each BS. In the existing literature, the centralized MARL subframe configuration requires the states of all BSs, which may not be easy to implement in practice due to the curse of dimensionality and privacy issues. Moreover, the decentralized MARL subframe configuration may not efficiently avoid inter-cell interference  if the BSs' learning processes are independent.
To tackle this challenge, inspired by the advantages of federated learning (FL) \cite{Blockchain2022Li}\cite{User-Level2022Wei}, we propose a federated reinforcement learning algorithm to design the dynamic resource allocation, aiming to meet heterogeneous UE demands while coordinating the inter-cell interference in a decentralized and privacy-preserving manner.


In this work, we propose a user-centric learning-based resource allocation framework in a heterogeneous cellular network consisting of multiple BSs, ground UEs (GUEs) and unmanned aerial vehicles (UAVs), where the BSs adaptively allocate time-frequency resources to satisfy the heterogeneous QoS demands characterized by the packet dropping ratio constraints. We summarize the main contributions as follows.
\begin{itemize}
\item We propose a learning-based D-TFDD scheme in a heterogeneous cellular system with dynamic UL and DL packet queuing processes. The proposed scheme exploits the merits of both D-TDD and  dynamic frequency division duplexing (D-FDD) by jointly allocating the time-frequency resources. We adopt D-TDD to adapt the BSs' subframe allocation to the asymmetric UL/DL traffic from a cell-centric perspective, and utilize D-FDD to cater the subchannel allocation to the heterogeneous QoS demands from a user-centric perspective.
\item We formulate the dynamic resource allocation problem under the proposed D-TFDD scheme as a decentralized partially observable MDP (Dec-POMDP), where each BS only has partial observation of the network environment. The BSs adaptively decide the subframe and subchannel allocations to maximize the long-term expected sum rate of the network while satisfying the UEs' packet dropping ratio constraints.
\item We propose a federated reinforcement learning algorithm named federated Wolpertinger deep deterministic policy gradient (FWDDPG) to solve the above optimization problem. The dimensionality of action space for D-TFDD control at each BS increases substantially as the number of UEs, subframes and subchannels increases. To deal with the large-scale discrete action space, we first adopt a DDPG-based policy at each BS to generate actions in a continuous space, and then discretize the actions based on Wolpertinger policy to reduce the mapping errors. For model aggregation across the BSs, we adopt a peer-to-peer FL architecture without a centralized server, where the BSs exchange their neural network  parameters with their one-hop neighbors to avoid privacy leakage and single point failure.
\item Simulation results show that our proposed D-TFDD scheme outperforms other benchmark TDD schemes, verifying the advantages of dynamically allocating multi-domain resources in serving heterogeneous UEs. Furthermore, the proposed algorithm outperforms independent DDPG (IDDPG) and it is even superior to the centralized multi-agent DDPG (MADDPG) by properly adjusting the Wolpertinger coefficient. The simulation reveals that, with sufficient system resources, the BSs prefer allocating more subchannels to the UEs with heavier traffic loads for throughput enhancement, and adopting similar subframe configurations across the cells for interference alleviation. Furthermore, in a resource-constrained regime, the BSs prioritize meeting local QoS constraints over avoiding interference.

\end{itemize}

The rest of this paper is organized as follows. In Section \uppercase\expandafter{\romannumeral2}, we present the system model of the proposed D-TFDD network. In Section \uppercase\expandafter{\romannumeral3}, we formulate the dynamic resource optimization problem as a Dec-POMDP. In Section \uppercase\expandafter{\romannumeral4}, we propose the FWDDPG algorithm to obtain the optimal resource allocation policies. Section \uppercase\expandafter{\romannumeral5} discusses the simulation results. At last, Section \uppercase\expandafter{\romannumeral6} concludes the paper.

\section{System Model}
We consider a heterogeneous multi-cell network which consists of a set ${\cal B} \triangleq \left\{ {1,2, \ldots ,B} \right\}$ of BSs serving a set ${\cal U} \triangleq \left\{ {{\cal U}_{\rm {GUE}} ,{\cal U}_{\rm {UAV}}} \right\}$ of UEs, where ${\cal U}_{\rm {GUE}}$ and ${\cal U}_{\rm {UAV}}$ denote the set of GUEs and UAVs, respectively. We denote ${{\cal U}^{b}}$ as the set of UEs served by BS $b$ inside cell $b$.

\begin{figure}[htb]
\centering
\label{Fig. 1}
\includegraphics[width=3.48 in]{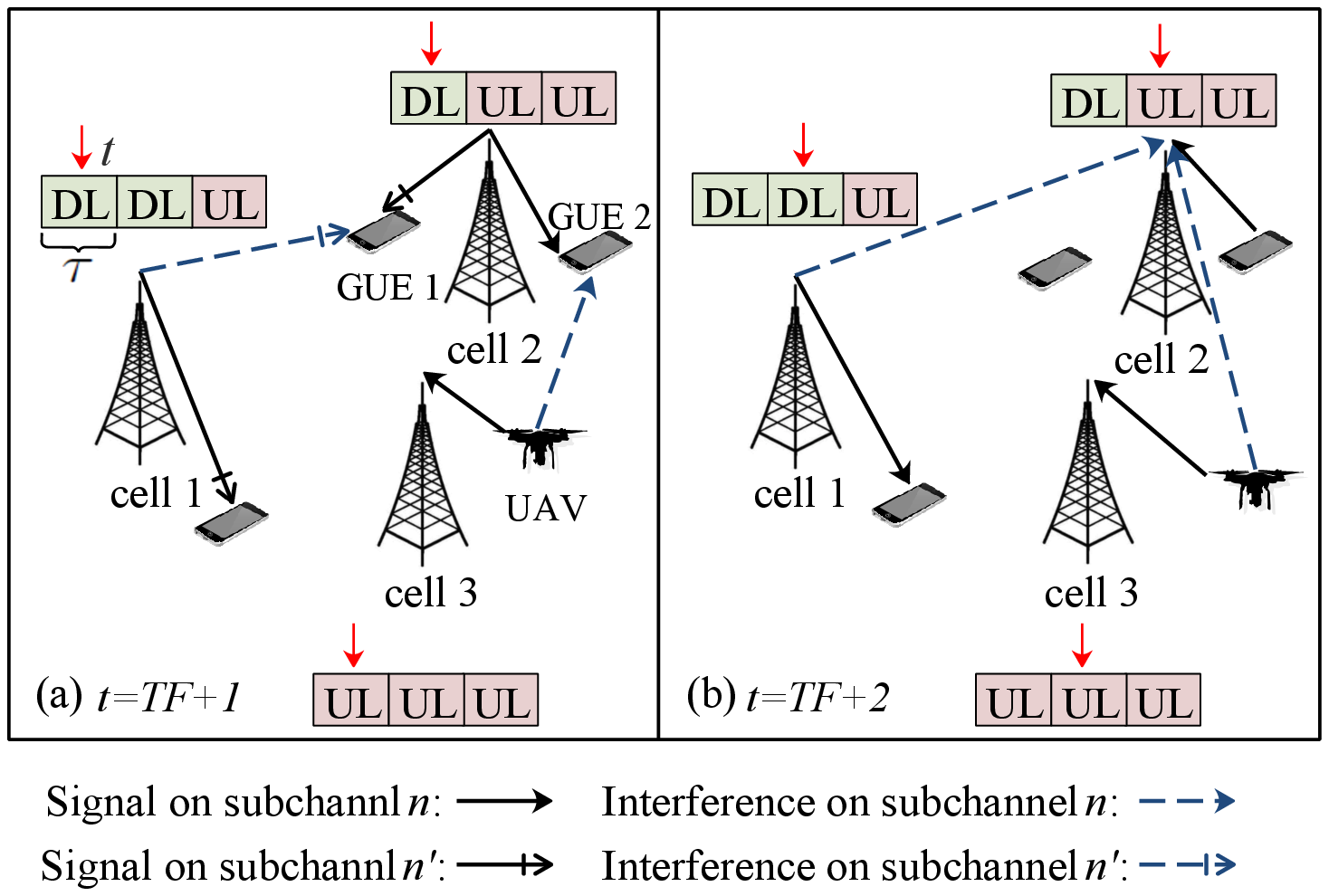}
\caption{ An illustration of proposed D-TFDD framework in time frame $T$. }
\end{figure}

A time-framed D-TFDD framework is shown in Fig.~1, where the DL/UL subframe configurations are dynamic across cells and time frames. Each time frame is made up of $F$ subframes, and the length of the subframe is $\tau$. For cell $b$ in time frame $T$, the first ${f ^b\left(T\right)}\in \left\{ {0,1, \ldots, F } \right\}$ number of successive subframes are used for DL transmissions and the rest of $F-f^b(T)$ number of successive subframes are used for UL transmissions.

We adopt orthogonal frequency division multiple access (OFDMA) for multiple access inside each cell, where the set of orthogonal subchannels is denoted by ${\cal N} \triangleq \left\{ {1,2, \ldots ,N} \right\}$ with the bandwidth $W$ for each subchannel. Assume that the number of subchannels is not less than the number of UEs served by any BS, i.e., $N\geq|{\cal U}^b|, \forall b$. Let ${\phi_{b,u}^{{n}}}\left(T\right) \in \left\{ {{\rm{0}},{\rm{1}}} \right\}$ denote whether or not subchannel $n$ is allocated to UE $u$ inside cell $b$ for DL transmissions, where ${\phi_{b,u}^{{n}}}\left(T\right) = 1$ denotes the subchannel $n$ is allocated to UE $u$ for $f^b\left(T\right)$ number of  successive DL subframes and ${\phi_{b,u}^{{n}}}\left(T\right) = 0$ means not. Similarly, for UL transmissions, the subchannel allocation is defined as ${\phi_{u,b}^{{n}}}\left(T\right) \in \left\{ {{\rm{0}},{\rm{1}}} \right\}$. Assume each subchannel $n$ can serve at most one receiver within a cell at a time, which can be represented as $\sum\nolimits_{u \in {{\cal U}^{b}}} {\phi_{b,u}^{{n}}}\left(T\right)\leq1$ and $\sum\nolimits_{u \in {{\cal U}^{b}}} {\phi_{u,b}^{{n}}}\left(T\right)\leq1$.

We consider quasi-static fading, where the channel state stays constant during each time frame for any given subchannel. Let $g_{\mathrm{tx},\mathrm{rx}}^n\left(T\right)$ denote the channel fading gain from transmitter $\mathrm{tx}$ to receiver $\mathrm{rx}$ on subchannel $n$ at time frame $T$, where $\mathrm{tx}$ and $\mathrm{rx}$ can be any UE $u \in {\cal U}$ or any BS $b \in {\cal B}$. The channel fading gain of $g_{\mathrm{tx},\mathrm{rx}}^n\left(T\right)$ includes both large-scale and small-scale fading \cite{AdaptiveFB2019Lee}. To compute the large-scale fading, the distance from transmitter $\mathrm{tx}$ to receiver $\mathrm{rx}$ is needed. We assume each UAV follows a pre-defined trajectory inside its associated cell (to fulfill specific tasks, e.g., surveillance), and the GUEs' and BSs' locations are static. For the ease of analysis, we discretize the flight trajectory of each UAV by a series of discrete locations, where we assume its location is static within a time frame $T$ and can be different across time frames \cite{Multi2020Cui}\cite{UAV2021Xiong}. Here, we adopt three-dimensional Cartesian coordinate and define the locations of transmitter $\mathrm{tx}$ and receiver $\mathrm{rx}$ at time frame $T$ as  $\left( {X_\mathrm{tx}}\left(T\right),{Y_\mathrm{tx}}\left(T\right),{H_\mathrm{tx}}\left(T\right) \right)$ and $\left( {X_\mathrm{rx}}\left(T\right),{Y_\mathrm{rx}}\left(T\right),{H_\mathrm{rx}}\left(T\right) \right)$, respectively. Then, the three-dimensional distance between transmitter $\mathrm{tx}$ and receiver $\mathrm{rx}$ is
\begin{equation}
\begin{split}
{\beta _{{\rm{tx}},\mathrm{rx}}}\left( T \right) =& \left\Vert {\left( {{X_{{\rm{tx}}}}\left( T \right),{Y_{{\rm{tx}}}}\left( T \right),{H_{{\rm{tx}}}}\left( T \right)} \right)} \right.\\
&{\left. { - \left( {{X_{{\rm{rx}}}}\left( T \right),{Y_{{\rm{rx}}}}\left( T \right),{H_{{\rm{rx}}}}\left( T \right)} \right)} \right\Vert_2},
\end{split}
\end{equation}
where $\Vert \cdot \Vert_2$ is Euclidean distance.  We adopt a general path loss model $\xi \left( \beta _{\rm{tx,rx}}\left( T \right)  \right)$ to consider both line-of-sight (LoS) and none-line-of-sight (NLoS) links. According to the well known International Telecommunication Union (ITU) model \cite{UAV2020Azari}\cite{Propagation2012ITU}, the probability of having a LoS link between transmitter $\mathrm{tx}$ and receiver $\mathrm{rx}$ is given by
\begin{equation}
\resizebox{0.91\hsize}{!}
{$\begin{aligned}
&{{\Pr} ^{\rm{LoS}}}\left( {\beta _{\rm{tx,rx}}\left( T \right)} \right) \\
&= \prod\limits_{j = 0}^{{c_4}} {\left[ {1 - \exp \left( { - {\textstyle{\frac{{{\left[ {{H_\mathrm{tx}}\left( T \right) - {\textstyle{\frac{\left( {j + 0.5} \right)\left( {{H_\mathrm{tx}}\left( T \right) - {H_\mathrm{rx}}\left( T \right)} \right)} {{c_4} + 1}}}} \right]}^2}} {{{\left( {\sqrt 2 {c_3}} \right)}^2}}}}} \right)} \right]},
\end{aligned}$}
\end{equation}
where $\left\{ {{c_1},{c_2},{c_3}} \right\}$ are environment-dependent parameters and ${c_4} = \left\lfloor {\frac{{\beta _{\rm{tx,rx}}\left( T \right)\sqrt {{c_1}{c_2}} }}{{1000}} - 1} \right\rfloor $. The probability of having a NLoS link between transmitter $\mathrm{tx}$ and receiver $\mathrm{rx}$ is given by
\begin{equation}
 {{\Pr} ^{\rm{NLoS}}}\left( {\beta _{\rm{tx,rx}}\left( T \right)} \right)  = 1-{{\Pr} ^{\rm{LoS}}}\left( {\beta _{\rm{tx,rx}}\left( T \right)} \right).
\end{equation}
Then, the general path loss model $\xi \left( \beta _{\rm{tx,rx}}\left( T \right)  \right)$ is given by
\begin{equation}
\resizebox{0.91\hsize}{!}
{$\begin{aligned}
\xi \left( {\beta _{\rm{tx,rx}}\left( T \right)} \right)
=\left\{ {\begin{array}{*{20}{l}}
{{A^{\rm{LoS}}}\beta _{\rm{tx,rx}}{{\left( T \right)}^{{\alpha ^{\rm{LoS}}}}},}&{{\rm{with \ prob.}} \ (2),}\\
{{A^{{\rm{NLoS}}}}\beta _{\rm{tx,rx}}{{\left( T \right)}^{{\alpha ^{{\rm{NLoS}}}}}},}&{{\rm{with \ prob.}} \ (3).}
\end{array}} \right.
\end{aligned}$}
\end{equation}
Let ${A^{\rm{LoS}}}$ and ${A^{\rm{NLoS}}}$ denote the reference path loss for LoS and NLoS links, and ${\alpha ^{\rm{LoS}}}$ and ${\alpha ^{\rm{NLoS}}}$ denote the path loss exponent for LoS and NLoS links, respectively. Furthermore, Nakagami-m small-scale fading is adopted in our model. Let $h_{\rm{tx,rx}}^n\left( T \right)$ denote the small-scale fading gain on subchannel $n$ between transmitter $\mathrm{tx}$ and receiver $\mathrm{rx}$ at time frame $T$, and the cumulative distribution function of $h_{\rm{tx,rx}}^n\left( T \right)$ can be obtained as
\begin{equation}
\begin{split}
\mathcal{F} \left( x \right)& \buildrel \Delta \over = \Pr \left[ {h_{{\rm{tx}},{\rm{rx}}}^n\left( T \right) \leq x} \right] \\
&= 1 - \sum\limits_{j = 0}^{{m_{{\rm{tx,rx}}}}} {\frac{{{{\left( {{m_{{\rm{tx,rx}}}}x} \right)}^j}}}{{j!}}} \exp \left( { - {m_{{\rm{tx,rx}}}}x} \right),
\end{split}
\end{equation}
where ${m_{{\rm{tx,rx}}}}$ is the fading parameter. Taking into account both the large-scale and small-scale fading, the channel fading gain is thus given by
\begin{equation}
g_{\rm{tx,rx}}^n\left( T \right) = \left[\xi {\left( {\beta_{\rm{tx,rx}} \left( T \right)} \right)}\right]^{ - 1}h_{\rm{tx,rx}}^n\left( T \right).
\end{equation}

Consider that a typical UE $u$ is associated with a typical BS $b$. Let ${P_{b}}$ and ${P_{u}}$ denote the transmit power of BS $b$ and UE $u$, respectively. Consider a DL receiver UE $u$ is receiving information from BS $b$ on subchannel $n$. UE $u$ may suffer from the BS-to-UE interference from the set of DL cells ${\cal B}^{n,{\rm{DL}}}\left(t\right)\backslash b$ and UE-to-UE interference from the set of UL cells ${\cal B}^{n,{\rm{UL}}}\left(t\right)$ in subframe $t$, where the total interference power received at UE $u$ is given by
\begin{equation}
\begin{split}
{I^{n}_{u}\left(t\right)} &= \sum\limits_{b' \in {\cal B}^{n,{\rm{DL}}}\left(t\right)\backslash b} {\sum\limits_{u' \in {\cal U}^{b'}} {\phi _{b',u'}^{n}\left(T\right){P_{b'}}g_{b',u}^{n}\left(T\right)}}\\
&\quad +\sum\limits_{b' \in {\cal B}^{n,{\rm{UL}}}\left(t\right)} {\sum\limits_{u' \in {\cal U}^{b'}} {\phi _{u',b'}^{n}\left(T\right){P_{u'}}g_{u',u}^{n}\left(T\right)}}.
\end{split}
\end{equation}
Note that, though we assume the channel fading gains remain unchanged during a time frame $T$, the set of interfering cells can be different across different subframes $t$ due to the dynamic time and frequency allocation. Therefore, the signal to interference plus noise ratio (SINR) at the DL receiver UE $u$ in subframe $t$ on subchannel $n$ is given by
\begin{equation}
{\rm{SINR}}_{b,u}^n\left(t\right) = \frac{{P_b}g_{b,u}^n\left(T\right)}{{I^n_{u}\left(t\right) + N_0W}},
\end{equation}
where $N_0$ is the variance of white Gaussian noise.
Consider that data transmission between any pair of transmitter $\mathrm{tx}$ and receiver $\mathrm{rx}$ is successful only if the received SINR is no less than a pre-defined threshold $\varsigma_\mathrm{rx}$. The DL achievable rate at UE $u$ is expressed as
\begin{equation}
R_{b,u}^{n}\left( t \right) = \mathbbm{1} \left( {{\rm{SINR}}_{b,u}^n\left( t \right)}\ge\varsigma_u \right)\tau W{\log _2}\left( {1 + \varsigma_u} \right),  \label{SINR UE threshold}
\end{equation}
where $\mathbbm{1} \left( \cdot \right)$ is the indicator function that takes the value of $1$ if the event happens and the value of $0$ if not.
Here, the achievable rate is measured in bits per subframe.
As the UE can operate on multiple subchannels, the total DL achievable rate $R_{b,u}\left( t \right)$ for UE $u$ is given by
\begin{equation}
R_{b,u}\left( t \right) = \sum\limits_{n \in {\cal N}} {\phi _{b,u}^{n}\left(T\right)R_{b,u}^{n}\left( t \right)}.
\end{equation}

Next, we discuss the UL achievable rate at a typical BS $b$. Consider a UL receiver BS $b$ that is operating on subchannel $n$ may receive the co-channel interference from adjacent DL and UL cells, i.e., the BS-to-BS interference from the set of DL cells ${\cal B}^{n,{\rm{DL}}}\left(t\right)$ and UE-to-BS interference  from the set of UL cells ${\cal B}^{n,{\rm{UL}}}\left(t\right)\backslash b$ in subframe $t$, which is expressed as
\begin{equation}
\begin{split}
I^n_{b}\left(t\right) &= \sum\limits_{b' \in {\cal B}^{n,{\rm{DL}}}\left(t\right)}{\sum\limits_{u' \in {\cal U}^{b'}} {\phi _{b',u'}^{n}\left(T\right){P_{b'}}g_{b',b}^{n}\left(T\right)}}\\
&\quad +\sum\limits_{b' \in {\cal B}^{n,{\rm{UL}}}\left(t\right)\backslash b} {\sum\limits_{u' \in {\cal U}^{b'}} {\phi _{u',b'}^{n}\left(T\right){P_{u'}}g_{u',b}^{n}\left(T\right)} }.
\end{split}
\end{equation}
The SINR and achievable rate at the UL receiver BS $b$ are respectively given by
\begin{equation}
{\rm{SINR}}_{u,b}^n \left(t\right)
= \frac{{P_u}g_{u,b}^n\left(T\right)}{{I^n_{b}\left(t\right) + {N_0W}}},
\end{equation}
and
\begin{equation}
R_{u,b}^{n}\left( t \right) = \mathbbm{1} \left( {{\rm{SINR}}_{u,b}^n\left( t \right)}\ge \varsigma_b \right)\tau W{\log _2}\left( {1 + \varsigma_b} \right). \label{SINR BS threshold}
\end{equation}
Therefore, for UE $u$, the total UL achievable rate $R_{u,b}\left( t \right)$ is given by
\begin{equation}
R_{u,b}\left( t \right) = \sum\limits_{n \in {\cal N}} {\phi _{u,b}^{n}\left(T\right)R_{u,b}^{n}\left( t \right)}.
\end{equation}

Each UE $u$ maintains a local UL queue and a DL queue at the BS side. At the beginning of time frame $T$ (before the data transmission), let $\hat{Q}_u^{\rm{DL}}(T)$ and $\hat{Q}_u^{\rm{UL}}(T)$ respectively denote the DL and UL queue lengths of UE $u$, which are the sizes of the  remaining packets in the DL and UL buffers.

For UE $u$ in time frame $T$, the amount of DL received packets at UE $u$ during $f^b\left(T\right)$ successive DL subframes is defined as
\begin{equation}
{\psi _u^{\rm{DL}}}\left( T \right)= {\rm{min}}\left\{ {\hat Q_{u}^{{\rm{DL}}}\left(T\right), \sum\limits_{t = TF + 1}^{TF + {f ^b\left(T\right)}} { R_{b,u}\left( t \right) }} \right\}. \label{DL_packets}
\end{equation}
where ${\psi _u^{\rm{DL}}}\left( T \right)$ cannot exceed the amount of packets in the current DL queue $\hat{Q}_u^{\rm{DL}}\left( T \right)$. Similarly, the amount of UL received packets at BS $b$ from UE $u$ during the remaining $F-f^b(T)$ subframes is given by
\begin{equation}
\resizebox{0.89\hsize}{!}
{$\begin{aligned}
{\psi _u^{\rm{UL}}}\left( T \right)= {\rm{min}}\left\{ {\hat Q_{u}^{{\rm{UL}}}\left(T\right), \sum\limits_{t = TF + 1 + {f ^b\left(T\right)}}^{\left( {T + 1} \right)F} { R_{u,b}\left( t \right) }} \right\}, \label{UL_packets}
\end{aligned}$}
\end{equation}
where ${\psi _u^{\rm{UL}}}\left( T \right)$ cannot exceed the amount of packets in the current UL queue $\hat{Q}_u^{\rm{UL}}\left( T \right)$.

For any UE $u$, we consider that UL and DL packets arrive at the end of each time frame $T$. Consider that the sizes of UL packets $D_u^{\mathrm{UL}}\left( T \right)$ and DL packets $D_u^{\mathrm{DL}}\left( T \right)$ follow Poisson processes of ${\cal P}\left( {\lambda ^{\rm{UL}}_{u}} \right)$ and ${\cal P}\left( {\lambda ^{\rm{DL}}_{u}} \right)$, respectively, where ${\lambda ^{\rm{UL}}_{u}}$ and $\lambda_u^{\rm{DL}}$ are the average UL and DL packet sizes of UE $u$, respectively.

Therefore, we can deduce that the DL queue length for UE $u$ at the beginning of time frame $T + 1$ evolves as
\begin{equation}
\hat Q_{u}^{{\rm{DL}}}\left(T+1\right)
=  {\hat Q_{u}^{{\rm{DL}}}\left(T\right) - {\psi^{{\rm{DL}}} _{u}}\left(T\right)}   +D^{\rm{DL}}_{u}\left(T\right),
\end{equation}
where DL buffer at the BS is assumed to be sufficiently large. Similarly, the UL queue length for UE $u$ evolves as
\begin{equation}
\resizebox{0.8\hsize}{!}
{$\begin{aligned}
&\hat Q_{u}^{{\rm{UL}}}\left(T+1\right) \\
&= {\rm{min}}\left\{ {{{\hat Q}^{\max}_u},{{{\hat Q_{u}^{{\rm{UL}}}\left(T\right) - \psi _{u}^{{\rm{UL}}}\left(T\right)} } } + D_{u}^{{\rm{UL}}}\left(T\right)} \right\},
\end{aligned}$}
\end{equation}
where $\hat{Q}_u^{\max}$ is UL data buffer size at UE $u$. Once the UL queue length exceeds the buffer size $\hat{Q}_u^{\max}$, the newly arrived packets will be dropped.

To characterize the reliability of UL transmission, we denote ${d_{u}\left(T\right)}$ as the dropping ratio of UE $u$ estimated  at the end of time frame $T$, which is the ratio of total dropped data to total arrived data over the most recent $T-\Gamma$ time frames, i.e.,
\begin{equation}
\setcounter{equation}{19}
\resizebox{0.89\hsize}{!}
{$\begin{aligned}
{d_{u}\left(T\right)}
= 1 - \frac{{\sum\limits_{l = \Gamma+1}^T {{\psi _u^{\rm{UL}}}\left( l \right)} }+{\hat Q_{u}^{{\rm{UL}}}\left(T+1\right)}-{\hat Q_{u}^{{\rm{UL}}}\left( \Gamma+1\right)}}{{\sum\limits_{l = \Gamma+1}^T  {{D^{\rm{UL}}_{u}}\left(l\right)} }},
\end{aligned}$}
\end{equation}
where $\Gamma  = \max \left[ {0,T - \Lambda } \right]$, and $\Lambda $ is the window size that removes the effect of the earlier history.


We consider that each BS can offer $E$ different types of slices, where each slice provides a customized service for the UEs with similar QoS requirements. Taking slice $e \in \left\{ {1, \cdots ,E} \right\}$ as an example, the set of UEs accessing slice $e$ is defined as ${\cal U}^e$, and the maximum tolerable dropping ratio for each UE in this slice is $d_e^{{\rm {max}}}$. The packet dropping ratio constraint is given by
\begin{equation}\label{task QoS}
  d_u\left(T\right)\leq d_e^{{\rm {max}}}.
\end{equation}
Our target is to joint optimize the subframe and subchannel allocation for maximizing the long-term sum rate under the UEs' packet dropping ratio constraints, i.e.,
\begin{equation}
\resizebox{0.89\hsize}{!}
{$\begin{aligned}
&\mathop {\max }\limits_{\left\{\phi _{b,u}^{n}\left(T\right),\phi _{u,b}^{n}\left(T\right),f^b\left(T\right),\forall b,\forall T\right\}} \mathop   \sum\limits_{T = 0}^\Psi  {\sum\limits_{u = 1}^U \left[{{\psi _u^{{\rm{DL}}}\left( T \right) + \psi _u^{{\rm{UL}}}\left( T \right)}} \right]},   \\
&\quad \quad \quad \quad \mathrm{s.t.} \quad\quad\quad \quad  d_u\left(T\right)\leq d_e^{{\rm {max}}}, \forall u,\forall T,
\end{aligned}$}
\end{equation}
where $\Psi$ is the total number of time frames.

\section{Decentralized Partially Observable MDP for D-TFDD Networks}
All the BSs coordinate to control the inter-cell interference and serve UEs in a decentralized way. Each BS independently makes the resource allocation decisions based on its local observations, with the aim of maximizing the long-term expected sum rate while satisfying the local QoS requirements of its serving UEs. We model this cooperative multi-agent task as a Dec-POMDP.

\emph{State:} Denote the joint state space of all BSs by ${\cal S} =  \otimes {{\cal S}^b}$, ${\forall b \in {\cal B}}$, with $\otimes$ as the Cartesian product, where ${\cal S}^b$ is the set of states of BS $b$. Considering that each BS only has partial observations of the network due to privacy issues. We denote the state of BS $b$ by
\begin{equation}
\setcounter{equation}{22}
{\mathbf{s}^b\left( T \right)} = \left\{  {\left. {\left( {{\hat Q}_{u}^{{\rm{UL}}}\left( T \right),{\hat Q}_{u}^{{\rm{DL}}}\left( T \right)} \right)} \right|u \in {\cal U}^b} \right\},
\end{equation}
which includes the current UL and DL queue lengths of all UEs served by this BS. The joint state of the network is denoted by $\mathbf{s}\left( T \right) =  \otimes {\mathbf{s}^b}\left( T \right)\in \mathcal{S}$.


\emph{Action:} Denote the action space of BS $b$ by ${{{\cal A}^b}}$ and the joint action space of all BSs by ${\cal A} =  \otimes {{\cal A}^b}$, ${\forall b \in {\cal B}}$. Let ${\mathbf{a}^b\left( T \right)}\in {\cal A}^b$ represent the action of BS $b$ in time frame $T$. Each BS's action is to decide the number of DL subframes $f^b\left(T\right)$, the DL subchannel allocation ${\phi_{b,u}^{{n}}}\left(T\right)$ and UL subchannel allocation ${\phi_{u,b}^{{n}}}\left(T\right)$, i.e.,
\begin{align}
{\bf{a}}^b\left( T \right) =& \left\{ {\left( {{{{f^b}\left( T \right)},{\left\{ {\phi _{b,u}^{n}\left( T \right)} \right\}}_{n \in \cal{N}}},} \right.} \right.\nonumber\\
&\left. {\left. {\left. {{\left\{ {\phi _{u,b}^{n}\left( T \right)} \right\}}_{n \in \cal{N}}} \right)} \right|u \in {{\cal{U}}^b}} \right\}.
\end{align}

\begin{remark}
We derive the size of action space for BS $b$ as follows. Take the UL subchannel allocation for BS $b$ as an example. Let $J$ denote the number of UEs that are allocated with at least one UL subchannels in this cell and ${{\eta _j}}$ denote the non-zero number of UL subchannels allocated to the $j$-th UE in this UE set. For each time frame, the UL subchannel allocation action has $\mathrm{Num}^{\mathrm{UL}}$ number of possible choices, i.e.,
\begin{equation}
\begin{split}\label{action number}
&\mathrm{Num}^{\rm{UL}}=\mathrm{Num}^{\rm{DL}}
=\sum\limits_{J = 0}^{|{\cal U}^b|} {\sum\limits_{{\eta _1=1}}^{N - J + 1} {\sum\limits_{{\eta _2=1}}^{N - J + 2 - {\eta _1}} { \cdots }}} \\
&\sum\limits_{{\eta _J=1}}^{N - \sum\nolimits_{j = 1}^{J - 1} {{\eta _j}} }C_{|{\cal U}^b|}^J {C_N^{{\eta _1}}C_{N - {\eta _1}}^{{\eta _2}}   \cdots   C_{N - \sum\nolimits_{j = 1}^{J - 1} {{\eta _j}} }^{{\eta _J}}},
\end{split}
\end{equation}
which is related to the total number of subchannels $N$ and UEs ${|{\cal U}^b|}$ served by BS $b$. First, BS $b$ selects $J\in\left\{0,\cdots,{|{\cal U}^b|}\right\}$ out of ${|{\cal U}^b|}$ UEs for subchannel assignment, which has $C_{|{\cal U}^b|}^J $ number of choices. Then, BS $b$ sequentially assigns the subchannels to these $J$ UEs, where the $j$-th UE can select from the remaining $N-\sum_{j'=1}^{j-1}\eta_{j'}$ subchannels and has $C^{\eta_j}_{N-\sum_{j'=1}^{j-1}\eta_{j'}} $ number of choices. We further denote the number of possible choices of DL subchannel allocation action by  $\mathrm{Num}^{\mathrm{DL}}$ and can easily deduce that $\mathrm{Num}^{\mathrm{DL}}=\mathrm{Num}^{\mathrm{UL}}$. Moreover, for each time frame, since the BS allocates the first $f^b\left( T \right)$ successive subframes for DL transmission, the subframe configuration action has $F+1$ number of possible choices. Therefore, the size of the action space $\left|{\cal{A}}^b\right|$ is given by
\begin{equation}
\setcounter{equation}{25}
\left|{\cal{A}}^b\right|=\mathrm{Num}^{\mathrm{UL}}\times\mathrm{Num}^{\mathrm{DL}}\times\left( F+1 \right),
\end{equation}
which increases rapidly with the number of subchannel $N$, the number of UE  $|{\cal U}^b|$ and the number of subframe $F$.\footnote{For example, when $N=5$,  $|{\cal U}^b|=3$ and $F=10$, we have $\left|{\cal{A}}^b\right|=11534336$.}
\end{remark}

Define policy of BS $b$ as a function mapping from the state space to action space, which is expressed as a conditional probability density function of
\begin{equation}
\resizebox{0.65\hsize}{!}
{$\begin{aligned}
&{\pi ^b}\left( {{{\bf{a}}^b}\left( T \right)\left| {{\bf{s}}^b}\left( T \right) \right.} \right)\\
&=\Pr \left( {{{\bf{A}}^b}\left( T \right) = {{\bf{a}}^b}\left( T \right)\left| {{{\bf{S}}^b}\left( T \right) = {{\bf{s}}^b}\left( T \right)} \right.} \right),
\end{aligned}$}
\end{equation}
where ${{\bf{S}}^b}\left( T \right)$ and ${{\bf{A}}^b}\left( T \right)$ denote the state and action of BS $b$ in time frame $T$ that have not yet been observed or taken, and ${{\bf{s}}^b}\left( T \right)$ and ${{\bf{a}}^b}\left( T \right)$ represent the observed state and executed action, respectively. We denote the policy profile of the BSs by
$\pi  = \left[ {{\pi ^1}, \ldots ,{\pi ^B}} \right]$.

\emph{Transition probability:} The joint action ${\bf{a}}\left( T \right) =  \otimes {\mathbf{a}^b}\left( T \right)\in \mathcal{A}$ causes the state transition of all BSs in time frame $T$. The transition probability $\rho$ of the entire network environment that moves from state ${{\bf{s}}}\left( T \right)$ to state ${{\bf{s}}}\left( T+1 \right)$ after taking joint action ${{\bf{a}}}\left( T \right)$ is assumed to be unknown by the BSs.

\emph{Reward:} Each BS receives an immediate reward $r^b(T)$ when action $\mathbf{a}^b(T)$ is executed in state $\mathbf{s}^b(T)$, i.e.,
\begin{equation}
\resizebox{0.89\hsize}{!}
{$\begin{aligned}
{r^b}\left( T \right) = \sum\limits_{u \in {{\cal U}^b}} {\left[ {\psi _u^{{\rm{DL}}}\left( T \right) + \psi _u^{{\rm{UL}}}\left( T \right)} \right.}
\left. { -  \mathbbm{1} \left( {{d_u}\left( T \right)> d_e^{\max }} \right)\varpi } \right], \label{reward}
\end{aligned}$}
\end{equation}
where $\psi _u^{{\rm{DL}}}\left( T \right)$ is the DL rate given in (\ref{DL_packets}), $\psi _u^{{\rm{UL}}}\left( T \right)$ is the UL rate given in (\ref{UL_packets}), and $\varpi$ is a positive constant that penalizes the violation of the QoS requirements. We assume that each BS can only observe its own reward as the reward is private information.

Due to the correlated queue dynamics and inter-cell interference, action $\mathbf{a}^b\left( T \right)$ affects not only the achievable rate and dropping ratio of BS $b$, but also that of other BSs in the subsequent time frames. We characterize the long-term sum-reward of all BSs in the cooperative system by $V\left( T \right)$, i.e.,
\begin{equation}
V\left( T \right) = \sum\limits_{l = T}^{\Psi} {\sum\limits_{b = 1}^B  {{\gamma ^{l-T}}{r^b}\left( {l } \right)} }, \label{return}
\end{equation}
where  $\gamma \in[0,1]$ is the discount factor that reflects the effect of future rewards.

Based on the above discussions, we define Dec-POMDP as a five-tuple of $\left( {\left\{ {{{\cal S}^b}} \right\}_{b \in {\cal B}}},{{\{ {{\cal A}^b}\} }_{b \in \cal B}},\rho,{{\{ {r^b}\} }_{b \in \cal B}},\gamma \right)$.
However, it is difficult to know the exact value of $V\left( T \right)$, due to the randomness of future states and actions. More specifically, the future states depend on the transition probability $\rho$, and the future actions depend on the joint policy $\pi$. Given joint action ${{{\bf{A}}}\left( T \right) = {{\bf{a}}}\left( T \right)}$ is taken at joint state $\mathbf{S}\left( T \right)={{\bf{s}}}\left( T \right)$, we define the conditional expectation of the long-term sum-reward of all BSs under joint policy $\pi$ as
\begin{align}\label{Q function}
&Q^{\pi}\left( {{{\bf{s}}}\left( T \right),{{\bf{a}}}\left( T \right)} \right) = \\ \nonumber
&\mathbb{E}_{{{\bf{S}}}\left( T+1 \right),{{\bf{A}}}\left( T+1 \right),\ldots}{\left[ {V\left( T \right)\left| {{{\bf{S}}}\left( T \right) = {{\bf{s}}}\left( T \right)}, {{{\bf{A}}}\left( T \right) = {{\bf{a}}}\left( T \right)} \right.} \right]},
\end{align}
which is also defined as the state-action value function. The objective of the BSs is to find the optimal joint policy $\pi^ {*} = \left[ {{\pi ^{1*}}, \ldots ,{\pi ^{B*}}} \right]$ that maximizes the state-action value function in (\ref{Q function}), i.e.,
\begin{equation}
\pi^ {*} = \mathop {\arg \max }\limits_{{\pi }} Q^{\pi}\left( {{{\bf{s}}}\left( T \right),{{\bf{a}}}\left( T \right)} \right), \forall{{\bf{s}}}\left( T \right),\forall{{\bf{a}}}\left( T \right).\label{optimization problem}
\end{equation}

\section{Federated Reinforcement Learning Based Resource Allocation Algorithm }
\begin{figure}[htb]
\centering
\includegraphics[width=3.4 in]{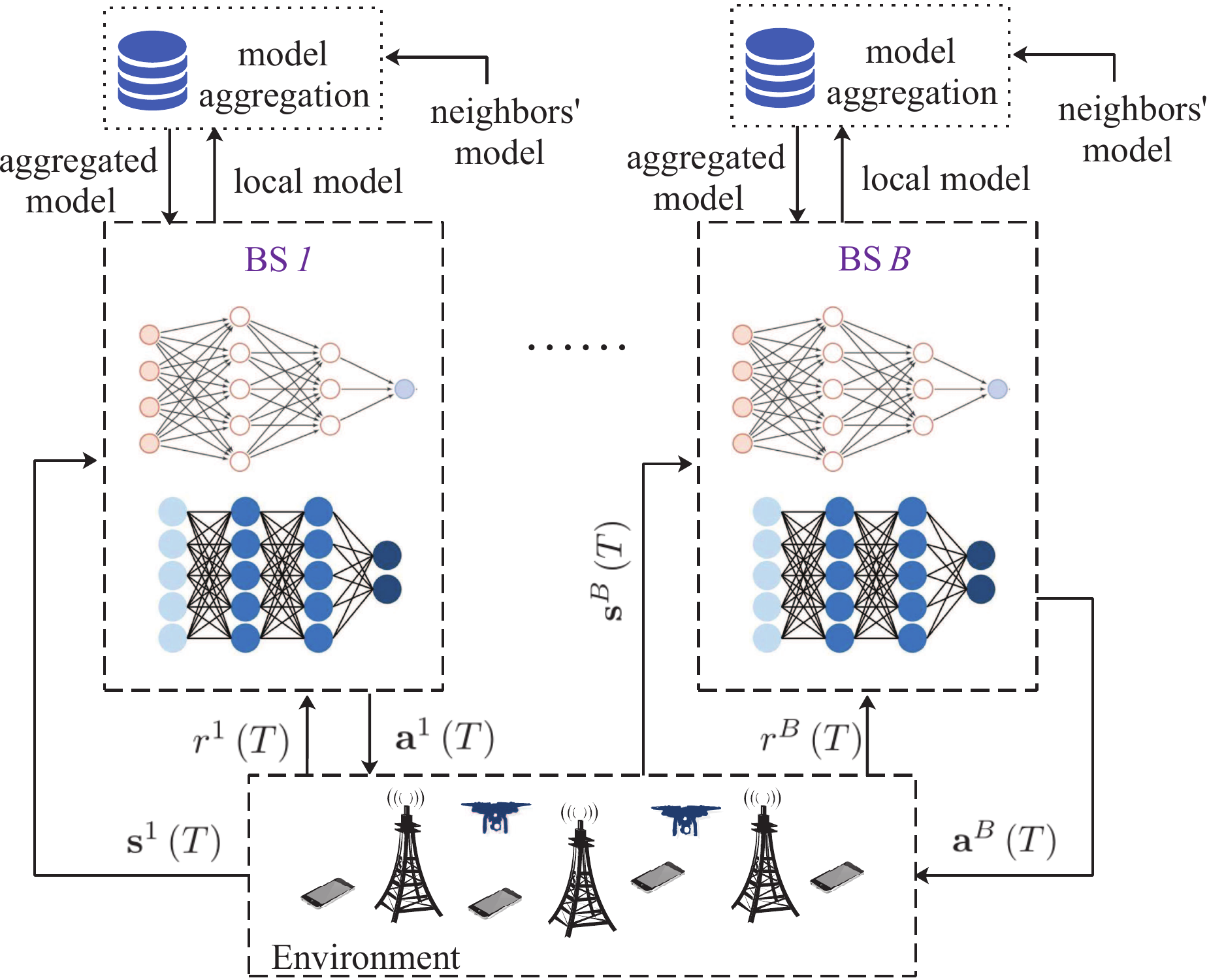}
\caption{ The architecture of the proposed FWDDPG algorithm. }
\label{Fig. 18}
\end{figure}

To solve the above resource allocation problem, there are two challenges to be addressed. The first challenge is to handle the large-scale discrete action space. As shown in \textit{Remark 1}, the dimensionality  of action space for BS $b$ is high when the numbers of subchannels, subframes and UEs are large. The conventional value-based reinforcement learning algorithms, e.g., deep Q network, may suffer from a long training convergence time  and is even not tractable due to the curse of dimensionality. The policy-based algorithms, e.g., DDPG, can deal with continuous action space and achieve good convergence \cite{Multihop2021Huang}\cite{Reconfigurable2020Huang}. To deal with the dimensionality of the large-scale discrete action space, we first adopt a DDPG-based algorithm to generate actions for each BS in the continuous action space, and then discretize the actions based on Wolpertinger policy  \cite{DeepReinforcementLI2015DulacArnold} to reduce the action mapping errors. Moreover,  the second challenge is to jointly optimize the sum-reward in a decentralized manner. The conventional MARL algorithms with centralized training, i.e., MADDPG \cite{multiagent2017Ryan}, suffer from the threats of privacy leakage since each BS is required to upload its local private information (e.g., states, actions and rewards) to the centralized controller for joint model training. To jointly optimize training among BSs, we propose a federated reinforcement learning algorithm named FWDDPG, where each BS performs local model training in a decentralized manner and updates the local model parameters by aggregating the parameters received from its one-hop neighbors.
The architecture of our proposed algorithm is shown in Fig. \ref{Fig. 18} and the
details will be described in the following subsections.
\subsection{Action Generation Based on Wolpertinger Policy}
We adopt an actor-critic based algorithm with a deterministic policy, i.e., DDPG, to deal with the high dimensional action space, where the policy maps from state to a deterministic action instead of a probability distribution over the actions. However, the actions generated by the deterministic policy are continuous and may not be within the action space of $\mathcal{A}^b$. To solve this problem, we further discretize the output of the actor network and adopt the Wolpertinger policy for mapping error reduction, which can be divided into two phases: action generation and action refinement.

\begin{figure*}[hb]
\newcounter{TempEqCnt}
\setcounter{TempEqCnt}{\value{equation}}
\setcounter{equation}{33}
\hrulefill
\begin{align}\label{actor1}
{\nabla _{\theta ^b}}J\left( {{{\theta ^b}}\left( T \right)} \right)
&=\frac{1}{I}{\sum\limits_{i}{\nabla _{\theta ^b}}Q\left( {{\bf{s}}^b}\left( i \right),{\mu}\left({{\bf{s}}^b}\left( i \right); {{{\theta ^b}}\left( T \right)},{\omega ^b}\left( T \right) \right);{\omega ^b}\left( T \right) \right) }.
\end{align}
\begin{align}
{\nabla _{\theta ^b}}J\left( {{{\theta ^b}}\left( T \right)} \right)
& \approx\frac{1}{I}{\sum\limits_{i} {\left( {{\nabla _{\theta ^b}}{\hat\mu}\left( {{\bf{s}}^b}\left( i \right);{{\theta^b}\left( T \right)} \right)} \right.}
\left. { \cdot {\nabla _{{{\hat {\bf{a}}}^b}}}Q\left( {{{\bf{s}}^b}\left( i \right),{{\hat {\bf{a}}}^b}\left( i \right);{\omega ^b}\left( T \right) \left| {{{\hat {\bf{a}}}^b}\left( i \right) = {\hat\mu}\left({{\bf{s}}^b}\left( i \right); {{{\theta ^b}}\left( T \right)} \right)} \right.} \right)} \right) }.\label{actor2}
\end{align}
\setcounter{equation}{36}
\begin{equation}\label{loss function}
\resizebox{0.93\hsize}{!}
{$\begin{aligned}
\mathrm{Loss}\left( {\omega ^b}\left( T \right) \right)
=\frac{1}{2I}{\sum\limits_{i} {{{\left[ {{Q }\left( {{{\bf{s}}^b}\left( i \right),{\bf{a}}^b\left( i \right);{\omega ^b}\left( T \right)} \right)-{r^b}\left( i \right) - \gamma {\tilde Q }\left( {{{\bf{s}}^b}\left( {i + 1} \right),\tilde\mu \left( \mathbf{s}^b \left( i+1 \right);\tilde\theta^b\left( T \right),{\tilde\omega ^b}\left( T \right) \right);{\tilde\omega ^b}\left( T \right)} \right)}
\right]}^2}}}.
\end{aligned}$}
\end{equation}
\begin{align}\label{loss}
{\nabla _{{\omega ^b}}}\mathrm{Loss}\left( {\omega ^b}\left( T \right) \right) &= \frac{1}{I}\sum\limits_{i}\left[\left( {{Q}\left( {{{\bf{s}}^b}\left( i \right),{{\bf{a}}^b}\left( i \right);{\omega ^b}\left( T \right)} \right) - {r^b}\left( i \right) - \gamma {{\tilde Q}}\left( {{{\bf{s}}^b}\left( {i + 1} \right),{\tilde\mu }\left( {{{\bf{s}}^b}\left( {i + 1} \right);{{\tilde \theta }^b}\left( T \right),{{\tilde \omega }^b}\left( T \right)} \right);{{\tilde \omega }^b}\left( T \right)} \right)} \right)\right.\\ \nonumber
 &\quad\left. \cdot {\nabla _{{\omega ^b}}}{Q}\left( {{{\bf{s}}^b}\left( i \right),{{\bf{a}}^b}\left( i \right);{\omega ^b}\left( T \right)} \right)\right]
\end{align}
\end{figure*}
\setcounter{equation}{\value{TempEqCnt}}


\emph{Action generation:} Given state ${\bf{s}}^b\left( T \right)$, the actor network of BS $b$ generates a proto-action $\hat{\mathbf{a}}^b(T)$ based on deterministic policy $\hat\mu$, i.e.,
\begin{equation}
\hat {\mathbf{a}}^b\left( T \right) = \hat \mu \left( {\mathbf{s}}^b\left( T \right);\theta^b\left( T \right) \right),
\end{equation}
where $\theta^b\left( T \right)$ is the neural network parameter to approximate policy $\hat \mu$ of BS $b$. However,  $\hat {\mathbf{a}}^b\left( T \right)$ is continuous and may not be a valid action in the discrete action set $\mathcal{A}^b$. Therefore, we map $\hat {\mathbf{a}}^b\left( T \right)$ to the elements of ${\cal A}^b$, i.e.,
\begin{equation}
{{\cal A}^b_{k}}\left( T \right)={\delta^b_k}\left( {\hat {\mathbf{a}}}^b\left( T \right) \right) = \mathop {\arg \min }\limits_{\mathbf{a}^b\left( T \right) \in {\cal A}^b}^k {\left\Vert {\mathbf{a}^b\left( T \right) - \hat {\mathbf{a}}^b\left( T \right)} \right\Vert_2}, \label{proto}
\end{equation}
where ${\delta^b_k}\left( {\hat {\mathbf{a}}}^b\left( T \right) \right)$ is the $k$-nearest-neighbor  ($k$-NN) mapping function to return the $k$ actions in ${\cal A}^b$ that are closest to $\hat {\mathbf{a}}^b\left( T \right)$ by Euclidean distance.


\emph{Action refinement:} We select the best action out of $k$ candidate actions generated by (\ref{proto}). The parameterized state-action value function of BS $b$ is defined as $Q\left( {{\mathbf{s}^b\left( T \right)},{\mathbf{a}^b\left( T \right)};{{ \omega }^b\left( T \right)}} \right)$, where $\omega^b\left( T \right)$ is the critic neural network parameter. To avoid picking an action with a low Q-value, we
adopt Wolpertinger policy, i.e.,
\begin{align}\label{qqq}
&\mu\left( {\mathbf{s}}^b\left( T \right);\theta^b\left( T \right),\omega^b\left( T \right) \right)\nonumber \\
&=\mathop {\arg \max }\limits_{\mathbf{a}^b\left( T \right)\in {{\cal A}^b_{k}}\left( T \right) }  {{Q}\left( \mathbf{s}^b\left( T \right),\mathbf{a}^b\left( T \right);{\omega ^b\left( T \right)} \right)}\nonumber\\
&={\mathbf{a}}^b\left( T \right)
\end{align}to refine the output of the critic network by selecting the action with the highest Q-value among the $k$-NN actions.
The Wolpertinger policy's algorithm is given in Algorithm \ref{algorithm 1}.

\begin{algorithm}[htpb]
 \caption{ Wolpertinger Policy for BS $b$}
 \begin{algorithmic}[1]\label{algorithm 1}
 \STATE Observe state $\mathbf{s}^b \left( T \right)$ from environment.
    \STATE {Receive proto-action within continuous action space based on the actor network:} $\hat {\mathbf{a}}^b\left( T \right) = \hat\mu\left( {\mathbf{s}^b}\left( T \right);{\theta^b}\left( T \right) \right)$.
    \STATE {Retrieve a set of $k$ approximately closest actions to $\hat {\mathbf{a}}^b\left( T \right)$:} ${{\cal A}^b_{k}}\left( T \right) = {\delta^b_k}\left( \hat     {\mathbf{a}}^b\left( T \right) \right)$.
    \STATE {Compute the action with the highest Q-value:} ${\mathbf{a}}^b\left( T \right)=\mu\left( {\mathbf{s}}^b\left( T \right);\theta^b\left( T \right),\omega^b\left( T \right) \right)$.
 \end{algorithmic}
\end{algorithm}

\begin{remark}
Note that the size $k$ of the generated action set is task specific. There is a tradeoff between  policy quality and computational cost. The policy quality can be evaluated by the difference between the highest Q-value achieved over all possible actions and the expected highest Q-value achieved by these $k$ closest actions \cite{DeepReinforcementLI2015DulacArnold}. It can be deduced that the policy quality increases with $k$.
Moreover, the additional computational complexity for Wolpertinger policy grows linearly with $k$,
where the details will be discussed in \textit{Remark 3} in the next subsection.
\end{remark}

\subsection{The Local WDDPG Policy Training}
In this subsection, we will discuss the training process of the actor and critic networks for WDDPG algorithm. We consider the model-free scenario with  no prior distribution of the network environment, and adopt the  conventional random strategies for initialization, i.e., randomly initialize critic and actor networks with parameters $\omega^b\left( 0 \right)$ and $\theta^b\left( 0 \right)$, $\forall b \in {\cal B}$, respectively.

In our proposed algorithm, we adopt off-policy, which involves two different policies of behavioral and target policies. We adopt Wolpertinger policy with Ornstein Uhlenbeck (OU) noise as the behavioral policy to encourage exploration, and use Wolpertinger policy without noise as the target policy. The learning data generated by behavioral policy is defined as a 4-element tuple $\left( {{\mathbf{s}^b\left( T \right)},{\mathbf{a}^b\left( T \right)},{r^b\left( T \right)},{\mathbf{s}^b\left( T+1 \right)}} \right)$ and is stored in the replay buffer (RB).
The target policy uses the samples stored in the RB to update itself.
With the experience replay and target networks, we next introduce the actor and critic network training processes.

\emph{Actor network training:} We define the target function of the actor network as the expectation of the parameterized state-action value function, i.e., $J\left( {{{\theta ^b}}\left( T \right)} \right)= \mathbb{E}_{{{\bf{S}}^b}\left( T \right)}\left[Q\left( {{\bf{S}}^b}\left( T \right),{\mu}\left({{\bf{S}}^b}\left( T \right); {{{\theta ^b}}\left( T \right)},{\omega ^b}\left( T \right) \right);{\omega ^b}\left( T \right) \right)\right]$. The expectation is taken over all possible values of unobserved state $\mathbf{S}^b\left( T \right)$ in time frame $T$ to remove the state randomness. To approximate the expectation over ${{{\bf{S}}^b}\left( T \right)}$, we  take a minibatch of $I$ transitions from RB, where the $i$-th transition is denoted by $\left( {\mathbf{s}^b\left( i \right),{\mathbf{a}^b\left( i \right)},r^b\left( i \right),\mathbf{s}^b\left( i+1 \right)} \right)$. We aim to find the optimal ${{{\theta ^b}}\left( T \right)}$ that maximizes $J\left( {{{\theta ^b}}\left( T \right)} \right)$ by adopting a deterministic policy gradient method, where the gradient ${\nabla _{\theta ^b}}J\left( {{{\theta ^b}}\left( T \right)} \right)$ can be derived in (\ref{actor1}) at the bottom of this page. However, as the action ${\mathbf{a}}^b\left( i \right)={\mu}\left({{\bf{s}}^b}\left( i \right); {{{\theta ^b}}\left( i \right)},{\omega ^b}\left( T \right) \right)$ executed by BS $b$ is discrete, the parameter $\theta^b \left( T \right)$ of the actor network can not be directly updated via deterministic policy gradient method. Therefore, we use the continuous proto-action ${{\hat {\bf{a}}}^b}\left( i \right)=\hat\mu\left( {\mathbf{s}}^b\left( i \right); {\theta^b}\left( T \right) \right)$ instead to derive the gradient of ${\nabla _{\theta ^b}}J\left( {{{\theta ^b}}\left( T \right)} \right)$ as given by (\ref{actor2}). Accordingly, the parameter $\theta^b \left( T+1 \right)$ is updated by
\begin{equation}\label{actor update}
\setcounter{equation}{36}
{\theta ^b}\left( {T + 1} \right)
\leftarrow {\theta ^b}\left( T \right) + {\beta ^b}{\nabla _{\theta ^b}}J\left( {{{\theta ^b}}\left( T \right)} \right),
\end{equation}
where ${\beta ^b}$ is the learning rate of actor network.

\emph{Critic network training:} We adopt temporal-difference (TD) learning to update $\omega^b \left( T \right)$. With the transition $\left( {\mathbf{s}^b\left( i \right),{\mathbf{a}^b\left( i \right)},r^b\left( i \right),\mathbf{s}^b\left( i+1 \right)} \right)$ sampled from RB, the estimated value called TD target is given by ${r^b}\left( i \right) + \gamma { Q }\left( {{{\bf{s}}^b}\left( {i + 1} \right),\mu \left( \mathbf{s}^b \left( i+1 \right);\theta^b\left( T \right),{\omega ^b}\left( T \right) \right);{\omega ^b}\left( T \right)} \right)$ and the output of the current critic network can be given by ${Q}\left( {{{\bf{s}}^b}\left( i \right),{{\bf{a}}^b}\left( i \right);{\omega ^b}\left( T \right)} \right)$. Note that bootstrapping occurs if we use the current critic network parameter ${\omega ^b}\left( T \right)$ for both the TD calculation and updating, which may cause a non-uniform overestimation of the optimal state-action value function. To avoid the bootstrapping and reduce the overestimation, we introduce the target actor and critic networks that are copied from the original actor and critic networks. Accordingly, the parameterized state-action value function of the target critic network of BS $b$ is denoted by $\tilde Q\left( {{\mathbf{s}^b\left( i+1 \right)},\tilde\mu \left( \mathbf{s}^b \left( i+1 \right);\tilde\theta^b\left( T \right),{{\tilde \omega }^b\left( T \right)} \right);{{\tilde \omega }^b\left( T \right)}} \right)$, where  $\tilde\mu \left( \mathbf{s}^b \left( i+1 \right);\tilde\theta^b\left( T \right),{{\tilde \omega }^b\left( T \right)} \right)$ is the target Wolpertinger policy, and $\tilde \theta^b \left( T \right)$ and $\tilde \omega^b \left( T \right)$ respectively denote the parameters of the target actor and critic networks. The TD target with respect to the target networks is given by ${r^b}\left( i \right) + \gamma {\tilde Q }\left( {{{\bf{s}}^b}\left( {i + 1} \right),\tilde\mu \left( \mathbf{s}^b \left( i+1 \right);\tilde\theta^b\left( T \right),{\tilde\omega ^b}\left( T \right) \right);{\tilde\omega ^b}\left( T \right)} \right)$. The loss function and its gradient are given by (\ref{loss function}) and (\ref{loss}), respectively. The parameter $\omega^b \left( T+1 \right)$ can be updated by
\begin{equation}
\setcounter{equation}{39}
 \omega^b \left( T+1 \right)\leftarrow\omega^b \left( T \right) + {{\bar{\beta}} ^b}{\nabla _{{\omega ^b}}}\mathrm{Loss}\left( {\omega ^b}\left( T \right) \right),\label{w}
\end{equation}
where ${{\bar{\beta}} ^b}$ is the learning rate of the critic network.
Moreover, the target critic and actor networks are updated every step with a small step size to confirm soft updating, i.e.,
\begin{equation}
{\tilde \omega^b \left( T+1 \right)} \leftarrow \kappa {\omega^b \left( T+1 \right)} + \left( {1 - \kappa } \right){\tilde \omega^b \left( T \right)}
\end{equation}
and
\begin{equation}
\tilde \theta^b \left( T+1 \right)  \leftarrow \kappa \theta^b \left( T+1 \right)  + \left( {1 - \kappa } \right)\tilde \theta^b \left( T \right),
\end{equation}
where $\kappa$ is the update step size.

\begin{figure*}[b]
\centering
\includegraphics[width=5.5 in]{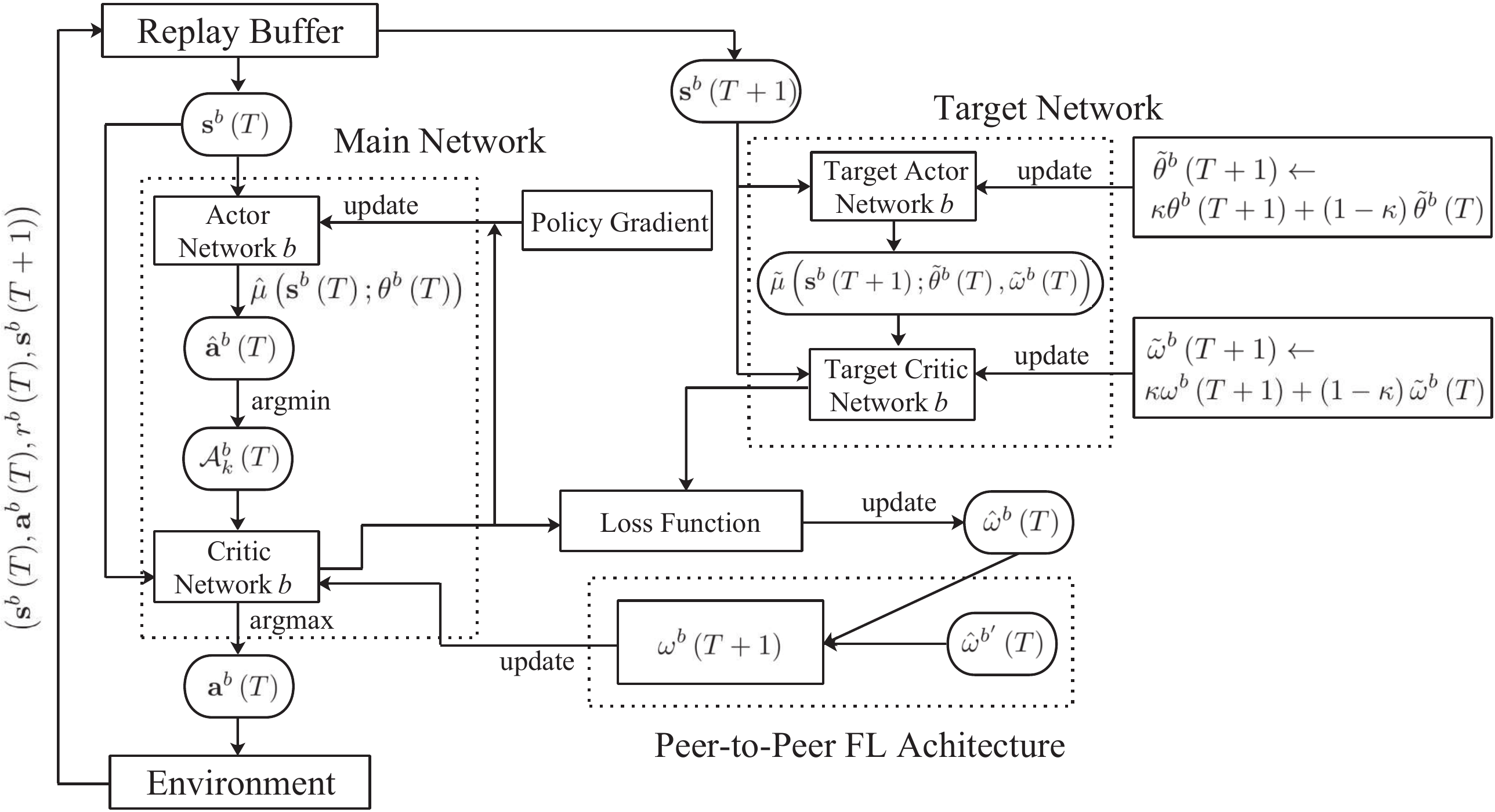}
\caption{ The framework of the proposed FWDDPG algorithm. }
\label{Fig. 4}
\end{figure*}

\begin{remark}
The computational complexity of WDDPG primarily depends on the actor and critic network architectures. Let $H_\mathrm{a}$ and $H_\mathrm{c}$ denote the total numbers of hidden layers of actor and critic networks. The $h$-th hidden layer for actor network and critic network involves $\zeta_{\mathrm{a},h}$ and $\zeta_{\mathrm{c},h}$ numbers of neurons, respectively. Recall that $|{\cal U}^b|$ denotes the number of UEs served by BS $b$. For the actor network, the number of neurons in the input layer depends on the dimension of the state, and the number of neurons in the output layer depends on the dimension of the action. Since the state of BS $b$ is defined as the current UL and DL queue lengths of its serving UEs, there are $2|{\cal U}^b|$ neurons in the input layer. And there are 3 neurons in the output layer corresponding to the three types of actions i.e., the number of DL subframe, DL and UL subchannel allocations. Accordingly, the number of weights in the input layer, the $h$-th ($2\leq h\leq H_\mathrm{a}-1$) hidden layer and the last hidden layer can be computed as $ 2|{\cal U}^b| \zeta_{\mathrm{a},1}$, $\zeta_{\mathrm{a},{h-1}}\zeta_{\mathrm{a},h}$ and $3\zeta_{\mathrm{a},H_\mathrm{a}}$, respectively. For the critic network, the number of neurons in the input layer is the dimension of the state and action, i.e., $2|{\cal U}^b|+3$, and there is 1 neuron in the output layer. Then the numbers of weights in the input layer, the $h$-th ($2\leq h\leq H_\mathrm{c}-1$) hidden layer and the last hidden layer can be computed as $\left( 2|{\cal U}^b|+3 \right)\zeta_{\mathrm{c},1}$, $\zeta_{\mathrm{c},{h-1}}\zeta_{\mathrm{c},h}$ and $\zeta_{\mathrm{c},H_\mathrm{c}}$, respectively. The computational complexity of BS $b$ in backward propagation training is given by ${\cal O}\left( \iota^{\mathrm{BP}} \left[  2|{\cal U}^b|\zeta_{\mathrm{a},1}+\sum_2^{H_\mathrm{a}}{\zeta_{\mathrm{a},{h-1}}\zeta_{\mathrm{a},h}}+3\zeta_{\mathrm{a},H_\mathrm{a}} + \left(2|{\cal U}^b|+3 \right)\right.  \right.$
$\left.\left.\times \zeta_{\mathrm{c},1} +\sum_2^{H_\mathrm{c}}{\zeta_{\mathrm{c},{h-1}}\zeta_{\mathrm{c},h}} + \zeta_{\mathrm{c},H_\mathrm{c}}
\right]  \right)$, where $\iota^{\mathrm{BP}}$ denotes the computational complexity for training a single weight in backward propagation. The computational complexity for training a single weight in forward propagation is similar to that in backward propagation. Here, we focus on the additional computational complexity caused by the Wolpertinger policy in forward propagation training, which is  given by ${\cal O}\left( \iota^{\mathrm{AP}} k \left[ \left(2|{\cal U}^b|+3 \right)\zeta_{\mathrm{c},1} +\sum_2^{H_\mathrm{c}}{\zeta_{\mathrm{c},{h-1}}\zeta_{\mathrm{c},h}} + \zeta_{\mathrm{c},H_\mathrm{c}}\right]  \right)$, where $\iota^{\mathrm{AP}}$ is the computational complexity of training a single weight in forward propagation.
\end{remark}

\subsection{Global Policy Training with Federated Learning}
Our objective is to find the optimal joint policy $\pi^ {*}$ that maximizes the global state-action value function in~(\ref{optimization problem}). The challenge is to maximize social welfare in a decentralized manner with local observations. If each BS independently adopts WDDPG algorithm, there is no communication overhead, but it suffers from low cooperation efficiency and can only adapt its resource allocation to the local traffic instead of the network. Due to the lack of global state information, it is difficult for the BSs to mitigate inter-cell interference among themselves. In order to alleviate inter-cell interference, it is necessary for the BSs to share local information with each other for joint model training. Although some conventional algorithms, e.g., MADDPG, can jointly train the critic networks at the centralized controller, each BS is required to upload its local states, actions, and rewards to the controller, which may cause privacy leakage issues and introduce high communication overhead. To protect privacy of the agents, we adopt a decentralized FL framework for joint model training among the BSs \cite{AResource2021Xue} \cite{Federated2021Qi}, where each BS exchanges the local critic network parameters with its one-hop neighbors every $\ell$ time frame. This enables the decentralized BSs to update their local critic network parameters to improve the global resource allocation efficiency with relatively low communication overhead. Note that our proposed scheme can indirectly exchange parameter information with multi-hop neighbors due to the propagation effect across multiple rounds of parameter update.

\begin{algorithm}[htpb]
 \caption{FWDDPG Based Resource Allocation Algorithm}
 \begin{algorithmic}[1]\label{algorithm 2}
 \STATE Randomly initialize critic and actor networks with parameters $\omega^b\left( 0 \right)$ and $\theta^b\left( 0 \right)$, $\forall b \in {\cal B}$.
 \STATE Initialize target critic and actor networks $\widetilde \omega^b\left( 0 \right) \leftarrow \omega^b\left( 0 \right)$, $\tilde \theta^b\left( 0 \right) \leftarrow \theta^b\left( 0 \right)$, $\forall b \in {\cal B}$.
 \STATE Initialize the $k$-NN mapping function $\delta^b_k$ using elements of ${\cal A}^b$, $\forall b \in {\cal B}$.
 \STATE Initialize RB.
  \STATE Initialize the number of subchannels $N$, the number of subframes $F$, the number of BSs $B$ and the number of UEs ${|{\cal U}^b|}$ served by BS $b$.
 \FOR{$\mathrm{Epoch}=1,2,\ldots  $}
 \STATE Initialize the global state $\mathbf{s}\left( 0 \right)$.
 \FOR{$T=0,1,2,\ldots  $}
 \FOR {$b=1$ to $B$}
 \STATE Observe local state $\mathbf{s}^b\left( T \right)$.
 \STATE Generate local action based on the Wolpertinger policy: $\mathbf{a}^b\left( T \right) = \mu\left( {\mathbf{s}}^b\left( T \right);\theta^b\left( T \right),\omega^b\left( T \right) \right)$.
 \ENDFOR
 \STATE Execute joint action ${\mathbf{a}}\left( T \right) = \left( {\mathbf{a}^1\left( T \right), \ldots ,\mathbf{a}^B\left( T \right)} \right)$.
 \FOR {$b=1$ to $B$}
 \STATE Observe  reward $r^b\left( T \right)$ and new  state ${\mathbf{s}^b\left( T+1 \right)}$.
 \STATE Store transition $\left( {\mathbf{s}^b\left( T \right),{\mathbf{a}^b\left( T \right)},r^b\left( T \right),\mathbf{s}^b\left( T+1 \right)} \right)$ in RB.
 \STATE Randomly sample a minibatch of $I$ transitions  from RB.
 \STATE Update the critic by minimizing the loss in (\ref{loss function}), then update ${{\hat \omega }^b}\left( T \right) \leftarrow {\omega ^b}\left( T \right)$.
 \STATE Update the actor using the sampled gradient according to (\ref{actor2}), then update $\theta^b \left( T+1 \right)\leftarrow \theta^b \left( T \right)$.
%
\STATE Update critic network according to (\ref{consus}).
 \STATE Update the target networks:

 ${\widetilde \omega^b \left( T+1 \right)} \leftarrow \kappa {\omega^b \left( T+1 \right)} + \left( {1 - \kappa } \right){\widetilde \omega^b \left( T \right)}$,

  $\tilde \theta^b \left( T+1 \right)  \leftarrow \kappa \theta^b \left( T+1 \right)  + \left( {1 - \kappa } \right)\tilde \theta ^b\left( T \right)$.
 \ENDFOR
 \ENDFOR
 \ENDFOR
 \end{algorithmic}
\end{algorithm}

We consider the D-TFDD network topology as a undirected graph model ${{\cal G}} = \left( {{\cal B},{\varrho }} \right)$, where ${\cal B}$ is the set of BS nodes and ${\varrho }$ represents the set of edges. An edge $\left( {b,b'} \right) \in {\varrho }$ means that BS $b'$ is the one-hop neighbor of BS $b$.
Let ${\Upsilon ^b} = \left\{ {b \in {\cal B}:\left( {b,b'} \right) \in {\varrho }} \right\}$ be the set of one-hop neighbors of BS $b$, where $| {{\Upsilon ^b}} |$ and $| {{\Upsilon ^{b'}}} |$ are the numbers of neighbors of BS $b$ and $b'$, respectively. Due to the differences in training capabilities and network connections of each neighboring BS $b'$, it is wise for BS $b$ to weight the model parameters received from its one-hop neighbors differently according to their influences. We denote the weighting matrix by ${Z} = {\left[ {{z_{b',b}}} \right]_{B \times B}}$, where $z_{b',b}$  weights the parameter sent from BS $b'$ to BS $b$. By adopting Metropolis weights \cite{Diffusion2008F.S.} in our model, we have
\begin{equation}
\setcounter{equation}{43}
\resizebox{0.89\hsize}{!}
{$\begin{aligned}
{z_{b',b}}
= \left\{ {\begin{array}{*{20}{l}}
 { \frac{1}{1 +{\max \left\{ {\left| {{\Upsilon ^b}} \right|,\left| {{\Upsilon ^{b'}}} \right|} \right\}}}} ,&{\forall \left( {b,b'} \right) \in \varrho },\\
{1 - \sum\limits_{b'' \in {\Upsilon  ^b}\left( T \right)} {{z_{b'',b}}} ,}&{b = b',\forall b \in {\cal B}}.
\end{array}} \right.
\end{aligned}$}
\end{equation}

For every $\ell$ time frames, each BS exchanges parameter $\hat \omega ^{b}\left( T \right)$ with its one-hop neighbors for global model training, where $\hat \omega ^{b}\left( T \right)=\omega^b \left( T \right) + {{\bar{\beta}} ^b}{\nabla _{{\omega ^b}}}\mathrm{Loss}\left( {\omega ^b}\left( T \right) \right)$. And then, each BS $b$ aggregates the received parameters $\hat \omega ^{b'}\left( T \right)$ from its one-hop neighbors based on the Metropolis weights and updates the parameter of the critic network in time frame $T+1$. For the rest of the time frames, BS $b$ directly uses its local parameter $\hat \omega ^{b}\left( T \right)$ to update its critic network. Therefore, the parameter update of the critic network can be expressed as
\begin{equation}
 \left\{ {\begin{array}{*{20}{l}}
\omega ^b\left( T+1 \right) \leftarrow\sum\limits_{b' =1}^{B} {{z_{b,b'}}\hat \omega ^{b'}\left( T \right)} ,&{\mathrm{if}\; T \%\ell =0 },\\
\omega ^b\left( T+1 \right) \leftarrow\hat \omega ^{b}\left( T \right),&{\mathrm{otherwise}}.
\end{array}} \right. \label{consus}
\end{equation}


We summarize the proposed FWDDPG algorithm  in Algorithm \ref{algorithm 2} and Fig. \ref{Fig. 4}.

\begin{remark}
The computational complexity  of the peer-to-peer FL architecture depends on the aggregation of critic network parameters from one-hop neighbors. For BS $b$, the critic network parameters of itself and its one-hop neighbors need to be multiplied by their respective weights and then added as the new critic network parameters for global training. We therefore can deduce that the computational complexity of the peer-to-peer FL architecture is ${\cal O}\left( \hbar\left[\sum_{b=1}^{B}\left(2\left| {{\Upsilon ^b}\left( T \right)} \right|+1\right)\right]\right)$, where $\hbar$ is the number of  rounds for global training, and the number of additions and multiplications are $\left| {{\Upsilon ^b}\left( T \right)} \right|$ and $\left| {{\Upsilon ^b}\left( T \right)} \right|+1$ for BS $b$, respectively.
\end{remark}
\section{Simulation Results and Discussions }
For simulations, we consider a D-TFDD network  covers a square area of $3$ km  $\times 3$ km. Without loss of generality, we consider ten BSs with the height of $10$ m serves $30$ active UEs (including GUEs and UAVs), where each BS serves three UEs in its serving area with $5$ subchannels and $10$ subframes. The transmit power of BSs and UEs are $24$ dBm and $23$ dBm, respectively. The noise power at BSs, GUEs and UAVs are $-91$ dBm, $-95$ dBm and $-99$ dBm, respectively \cite{Further20123GPP}\cite{Technical20173GPP}. As for the channel modeling, we set the ITU model factors $\left\{ c_1,c_2,c_3 \right\}$ as $\left\{0.3,500,20\right\}$ and the fading parameter $m_{\mathrm{tx},\mathrm{rx}}=1$ according to \cite{UAV2020Azari}. The parameters of the path loss model are listed in Table \uppercase\expandafter{\romannumeral1} according to \cite{Further20123GPP} and~\cite{UAV2020Azari}. The SINR threshold of UEs and BSs are set as $0$ dB and $-3$ dB, respectively, and the bandwidth of each subchannel is $10$ MHz. The duration of each subframe is $1$ ms. In the following discussions, we assume GUEs and UAVs are with slice types 1 and 2, respectively. Unless otherwise specified, the slice model parameters are given as follows. The maximum dropping ratio for GUEs and UAVs are set as $d_1^{\max}=0.3$ and $d_2^{\max}=0.1$, respectively. The average UL and DL packet sizes for GUEs are $\lambda_1^{\mathrm{UL}}=150$ KB and $\lambda_1^{\mathrm{DL}}=200$ KB, and those  for UAVs are $\lambda_2^{\mathrm{UL}}=50$ KB and $\lambda_2^{\mathrm{DL}}=80$ KB, respectively. The buffer sizes for GUEs and UAVs are $\hat Q^{\max}_1=250$ KB and $\hat Q^{\max}_2=150$ KB, respectively.
\begin{table}[htpb]
  \centering
   \setlength{\belowcaptionskip}{0.3 cm}
  \caption{The parameters of path loss model}
  \begin{tabular}{|c|c|}
    \hline
    Parameters& Values \\
    \hline
    \begin{tabular}[c]{@{}l@{}} BS-to-GUE \\ path loss factor \end{tabular}& \begin{tabular}[c]{@{}l@{}} $A^{\rm{L}}=34.02$ dB, $\alpha^{\rm{L}}=2.2$, \\$A^{\rm{NL}}=19.56$ dB, $\alpha^{\rm{NL}}=3.9$.  \end{tabular} \\
    \hline
    \begin{tabular}[c]{@{}l@{}} BS-to-UAV \\ path loss factor \end{tabular}& \begin{tabular}[c]{@{}l@{}} $A^{\rm{L}}=34.02$ dB, $A^{\rm{NL}}=20.96$ dB,\\ $\alpha^{\rm{L}}=2.2$,  $\alpha^{\rm{NL}}=4.6-0.7{\log _{10}}H_u\left( T \right)$.  \end{tabular} \\
    \hline
    \begin{tabular}[c]{@{}l@{}} BS-to-BS \\ path loss factor \end{tabular} & \begin{tabular}[c]{@{}l@{}} $A^{\rm{L}}=38.4$ dB, $\alpha^{\rm{L}}=2$, \\$A^{\rm{NL}}=49.36$ dB, $\alpha^{\rm{NL}}=4$. \end{tabular} \\
    \hline
    \begin{tabular}[c]{@{}l@{}} UAV-to-UAV \\ path loss factor \end{tabular}& \begin{tabular}[c]{@{}l@{}} $A^{\rm{L}}=34.02$ dB, $A^{\rm{NL}}=20.96$ dB,\\ $\alpha^{\rm{L}}=2.2$,  $\alpha^{\rm{NL}}=4.6-0.7{\log _{10}}H_u\left( T \right)$.  \end{tabular}  \\
    \hline
    \begin{tabular}[c]{@{}l@{}} GUE-to-GUE \\ path loss factor \end{tabular}&  \begin{tabular}[c]{@{}l@{}} $A^{\rm{L}}=38.4$ dB, $\alpha^{\rm{L}}=2$, \\$A^{\rm{NL}}=49.36$ dB, $\alpha^{\rm{NL}}=4$. \end{tabular}  \\
    \hline
  \end{tabular}
  \end{table}

%

The total number of  training epochs is 1000 and the number of  steps for each epoch is 300. We adopt two hidden layers for both actor and critic networks, where the first hidden layer has 60 neurons and the second hidden layer has 50 neurons. We train the neural networks by Adam optimizer, where we set the learning rates for the actor and critic networks as $0.0001$ and $0.001$, respectively.  For each time frame, a mini-batch of $300$ experiences are randomly sampled every time from RB that is capable of storing $1000000$ past experiences. We update the target critic or actor network by step size $\tau = 0.001$. We set the discount factor $\gamma = 0.99$. Unless otherwise specified, we adopt $k=120$ as the default size of the actions generated by Wolpertinger policy.

\begin{figure}[htb]
\centering
\includegraphics[width=3.5 in]{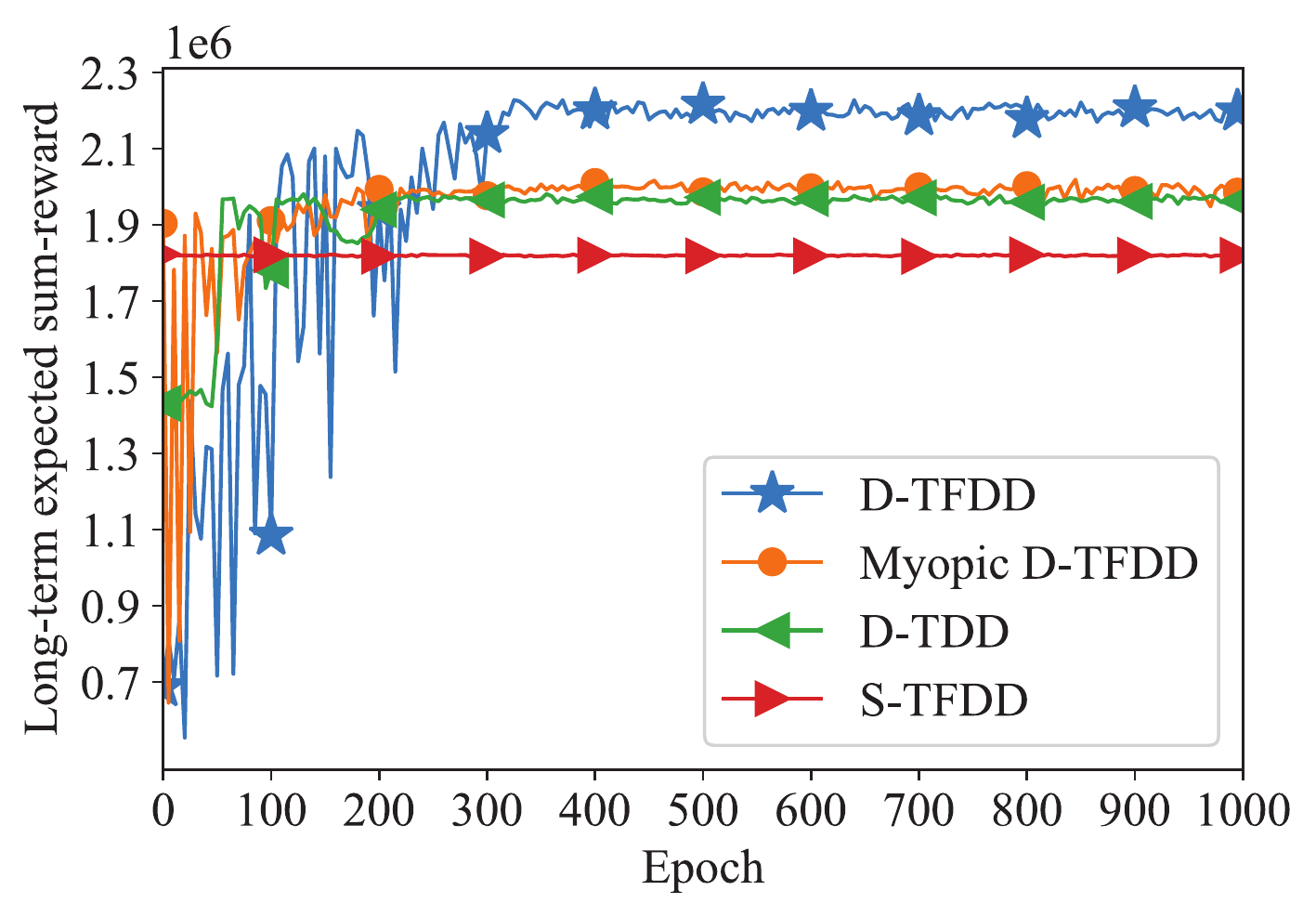}
\caption{ Long-term expected sum-reward of all BSs for different types of TDD schemes. }
\label{Fig. 15}
\end{figure}

In Fig. \ref{Fig. 15}, we plot the long-term expected sum-reward of the BSs over $1000$ training epochs. By adopting the proposed FWDDPG algorithm, we compare our D-TFDD scheme with other benchmark TDD schemes, i.e., S-TFDD, myopic D-TFDD and D-TDD. For our proposed D-TFDD scheme, both the subframe and subchannel allocations are adaptive to the UEs' dynamic traffic demands, aiming to maximize the long-term expected sum-reward of all BSs.
For static-TFDD (S-TFDD) scheme, all the BSs adopt the same subframe and subchannel configurations, which are pre-defined and non-adaptive throughout time. For myopic D-TFDD scheme, the subframe and subchannel configurations are adaptive to the UEs' dynamic demands in the current time frame only without considering the future rewards. For D-TDD scheme, only the subframe configuration is adaptive to the dynamic traffic demands, aiming at maximizing the long-term expected sum-reward of all BSs, while the subchannel allocation is pre-determined and does not change across time. In Fig. \ref{Fig. 15}, the performance of S-TFDD scheme does not change much over time and is worse than the dynamic schemes since it is not adaptive to the dynamic UE demands. We notice that there are slight jitters along the curve, which is due to the randomness of the channel gains and packet arrivals, although these effects are almost averaged out over the long-term accumulation. We also see that our proposed D-TFDD scheme outperforms all other benchmark schemes. It has better performance than myopic D-TFDD scheme since it considers not only the short-term but also the long-term sum-reward. Furthermore, it takes into account both the dynamic subframe and  subchannel allocations and is therefore better than D-TDD scheme.

\begin{figure}[htb]
\centering
\includegraphics[width=3.5 in]{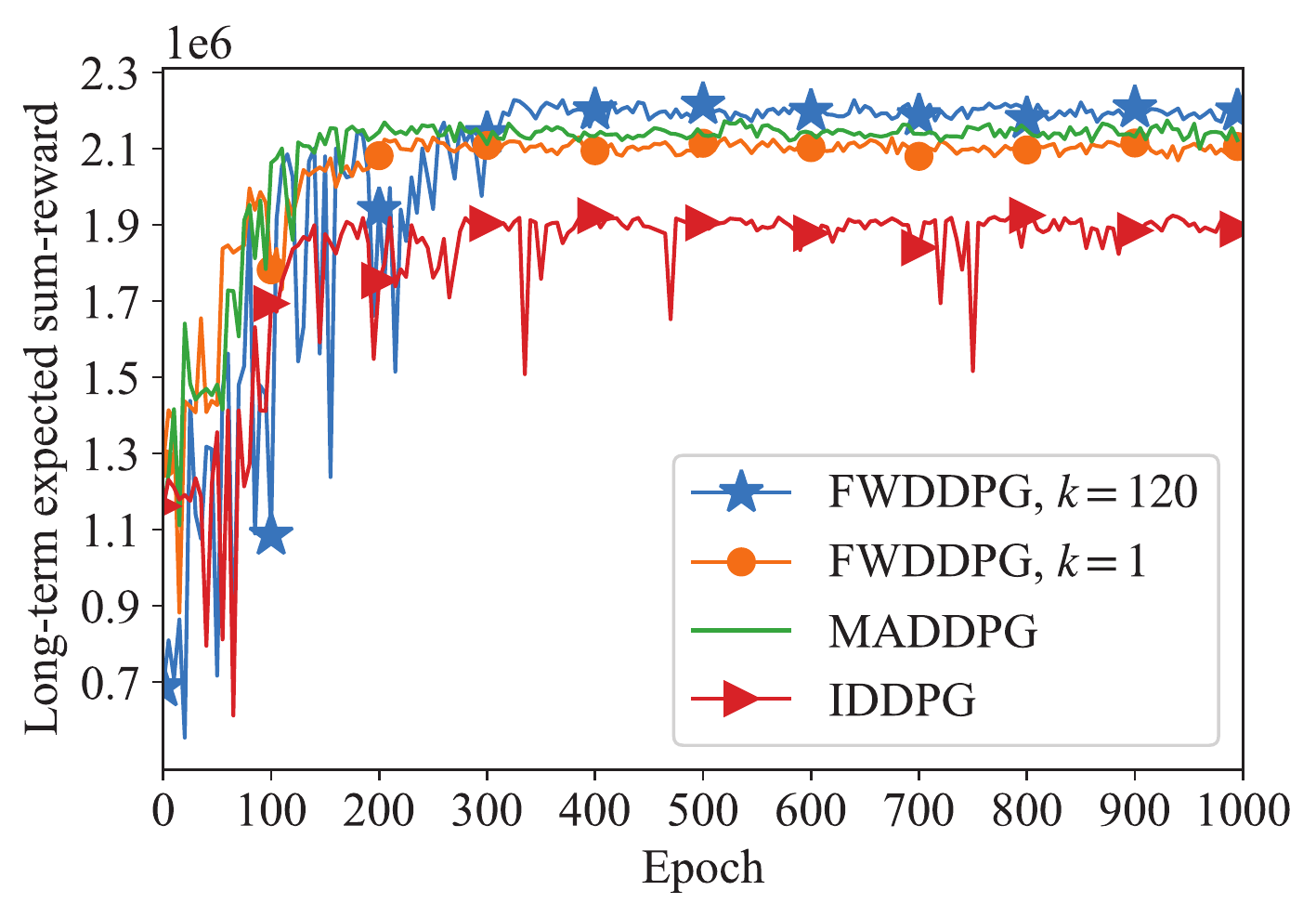}
\caption{ Long-term expected sum-reward of all BSs for D-TFDD scheme under different MARL algorithms. }
\label{Fig. 5}
\end{figure}

Fig. \ref{Fig. 5} depicts the long-term expected  sum-reward of the proposed FWDDPG algorithm and compares it with two benchmark algorithms, i.e.,  MADDPG \cite{multiagent2017Ryan} and IDDPG. For MADDPG algorithm, the centralized training and decentralized execution framework is adopted, where the BSs upload the local states, actions and rewards to the centralized controller to jointly train the critic network to maximize the long-term expected sum-reward of all BSs in the network. For IDDPG algorithm, each BS trains its DDPG algorithm with local states in a non-cooperative manner, aiming to maximize its local long-term expected reward \cite{QoE2019Tsai}~\cite{Dynamic2014Wang}. In Fig. \ref{Fig. 5}, we see that the sum-reward increases with the number of training epochs, which means all the algorithms can learn from interacting with the environment. Moreover, we see that IDDPG algorithm performs the worst among the three algorithms since the BSs are not cooperative. Next, we compare the performance of the proposed FWDDPG algorithm with  MADDPG algorithm. First, we observe that MADDPG algorithm outperforms FWDDPG algorithm with $k=1$. This is because MADDPG jointly trains the critic networks with the centralized controller, which is more efficient than the decentralized approaches. For $k=1$, only the discrete action that is closest to the continuous action is selected for execution, where the proposed algorithm is equivalent to that without Wolpertinger policy. However, this disadvantage can be compensated by adjusting the coefficient $k$ in the proposed FWDDPG algorithm. For example, for $k=120$, we see that the performance of FWDDPG algorithm exceeds that of MADDPG algorithm. Intuitively, this is because a larger $k$ can help include more candidates of valid actions, which increases the chance of selecting a better policy with a higher Q-value, though it may be at the cost of slower convergence speed.

\begin{figure}[htb]
\centering
\includegraphics[width=3.5 in]{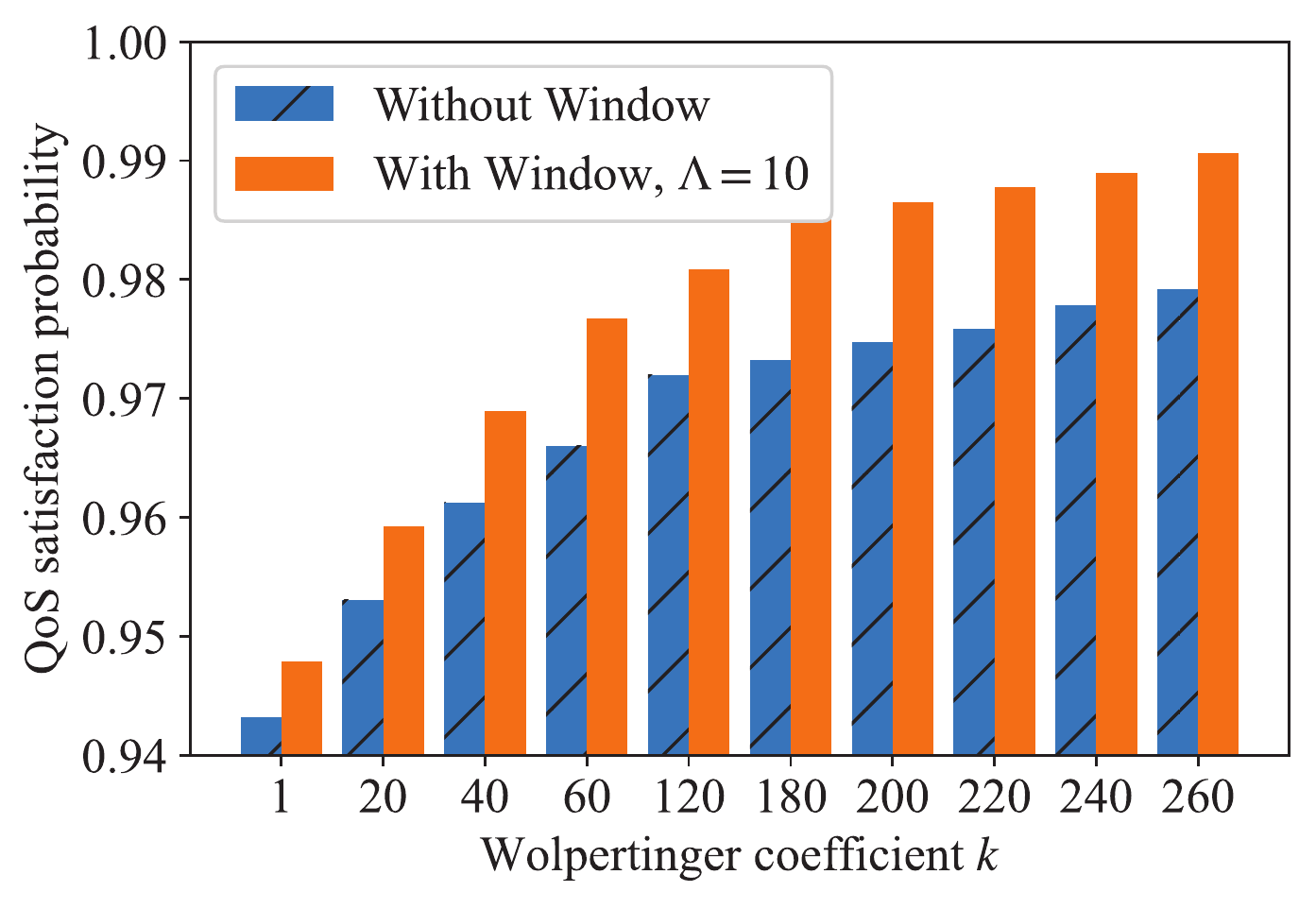}
\caption{ QoS satisfaction probability in the D-TFDD network versus various Wolpertinger coefficients $k$. }
\label{Fig. 6}
\end{figure}

In Fig. \ref{Fig. 6}, we plot the QoS satisfaction probability (the probability that the packet dropping ratio constraint is satisfied) in the D-TFDD network against the Wolpertinger coefficient $k$. On the one hand, we can see that the QoS satisfaction probability increases with $k$. This is because the policy quality improves as $k$ increases, which is consistent with \textit{Remark 2}. On the other hand, the computational complexity of WDDPG algorithm increases linearly with $k$ according to \textit{Remark 3}. We therefore can deduce that there exists an optimal value of $k$ that balances the policy quality and computational complexity. Furthermore, we observe that the QoS satisfaction probability is increased by introducing the sliding window in (19). If no sliding window is adopted, the premature experiences from the very first time frame will be included in the dropping ratio, which therefore reduces the QoS satisfaction probability. By using the sliding window, we can remove the effects of earlier history and thus improve the system performance.

\begin{figure}[htb]
\centering
\includegraphics[width=3.5 in]{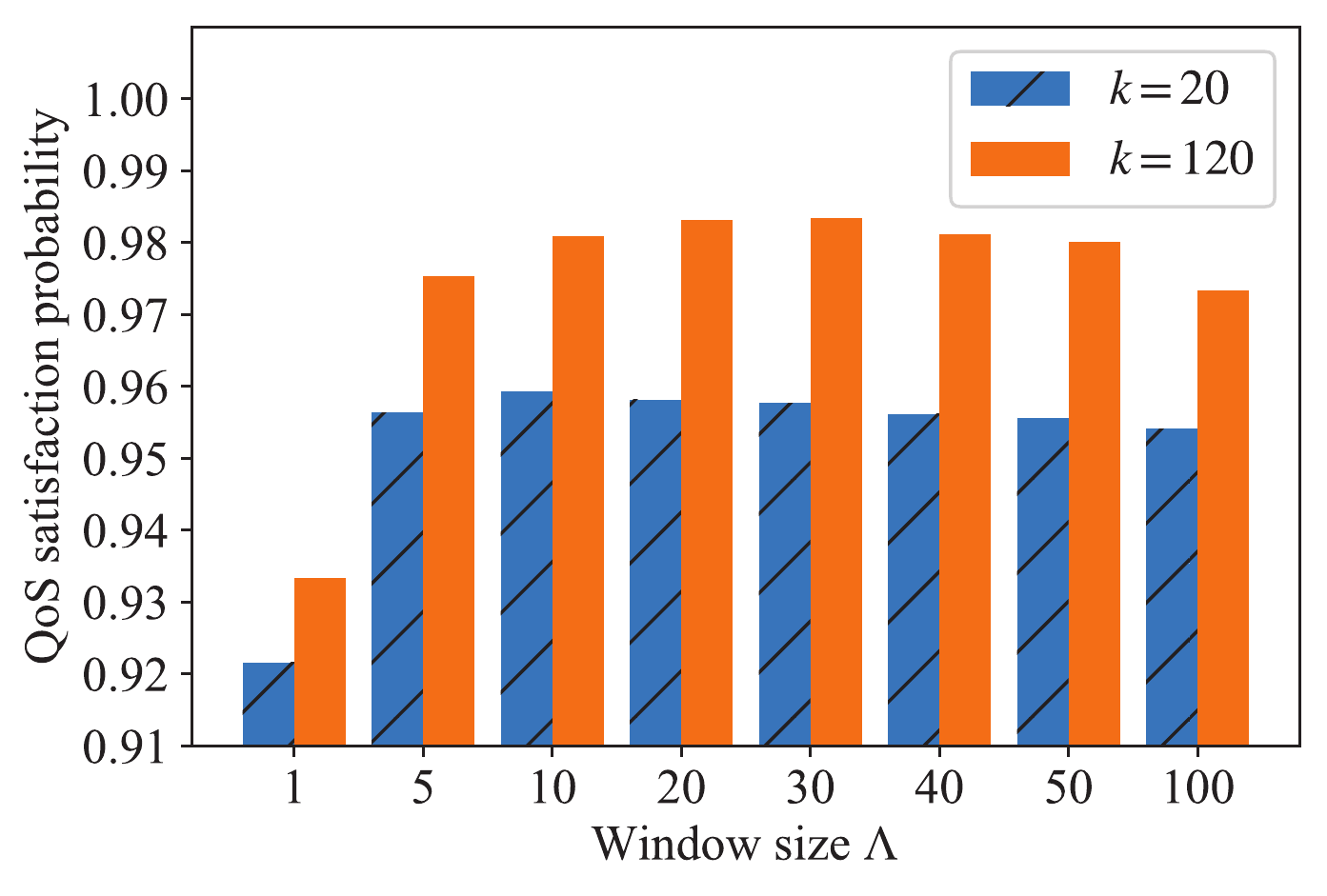}
\caption{QoS satisfaction probability versus various window sizes $\Lambda$. }
\label{Fig. 16}
\end{figure}

Furthermore, Fig. \ref{Fig. 16}  shows the influence of window size $\Lambda$ on the QoS satisfaction probability. We can see that the QoS satisfaction probability first increases and then decreases with the window size. When the window size is small, it means that only the latest samples are taken into the estimation of dropping ratio. The small number of samples leads to inaccurate representation of rewards, resulting in a low QoS satisfaction probability. When the window size increases, the increasing number of samples enhances the estimation accuracy of the dropping ratio and thus improves the QoS satisfaction probability. As window size further increases, more samples from the early history are included, which reduces QoS satisfaction probability.

\begin{figure}[htb]
\centering
\includegraphics[width=3.5 in]{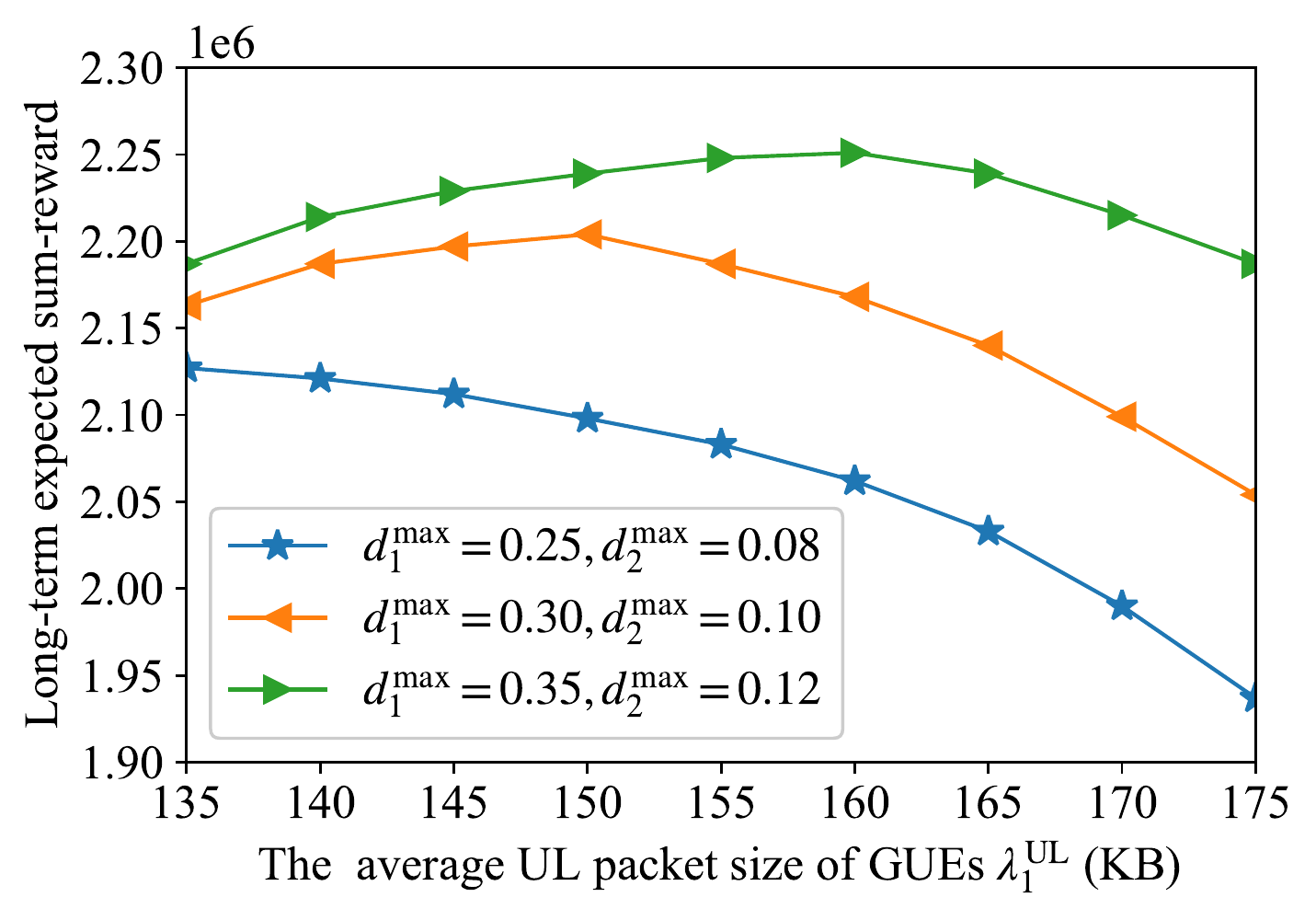}
\caption{ Long-term expected sum-reward of all BSs versus various average UL packet sizes ($\lambda_1^{\mathrm{DL}}=\lambda_1^{\mathrm{UL}}+50$ KB, $\lambda_2^{\mathrm{DL}}=\lambda_1^{\mathrm{UL}}-70$ KB, $\lambda_2^{\mathrm{UL}}=\lambda_1^{\mathrm{UL}}-100$ KB).}
\label{Fig. 11}
\end{figure}

Fig. 8 plots the influence of the average packet size and QoS constraint (i.e., maximum tolerable dropping ratio) on the long-term expected sum-reward of all BSs in the D-TFDD network. When the QoS constraint is not tight (e.g., $d_1^{\max}=\{0.30,0.35\}$, $d_2^{\max}=\{0.10,0.12\}$), the sum-reward first increases and then decreases with the average packet size. As the average packet size increases, the sum-reward first increases owning to the improvement in the sum rate. However, with the further increase of the average packet size, the sum-reward decreases due to the violation of the QoS constraints. Furthermore, when the QoS constraint is tight, the sum-reward  decreases directly with the average packet size because the QoS requirement is not met.

\begin{figure}[htb]
\centering
\includegraphics[width=3.9 in]{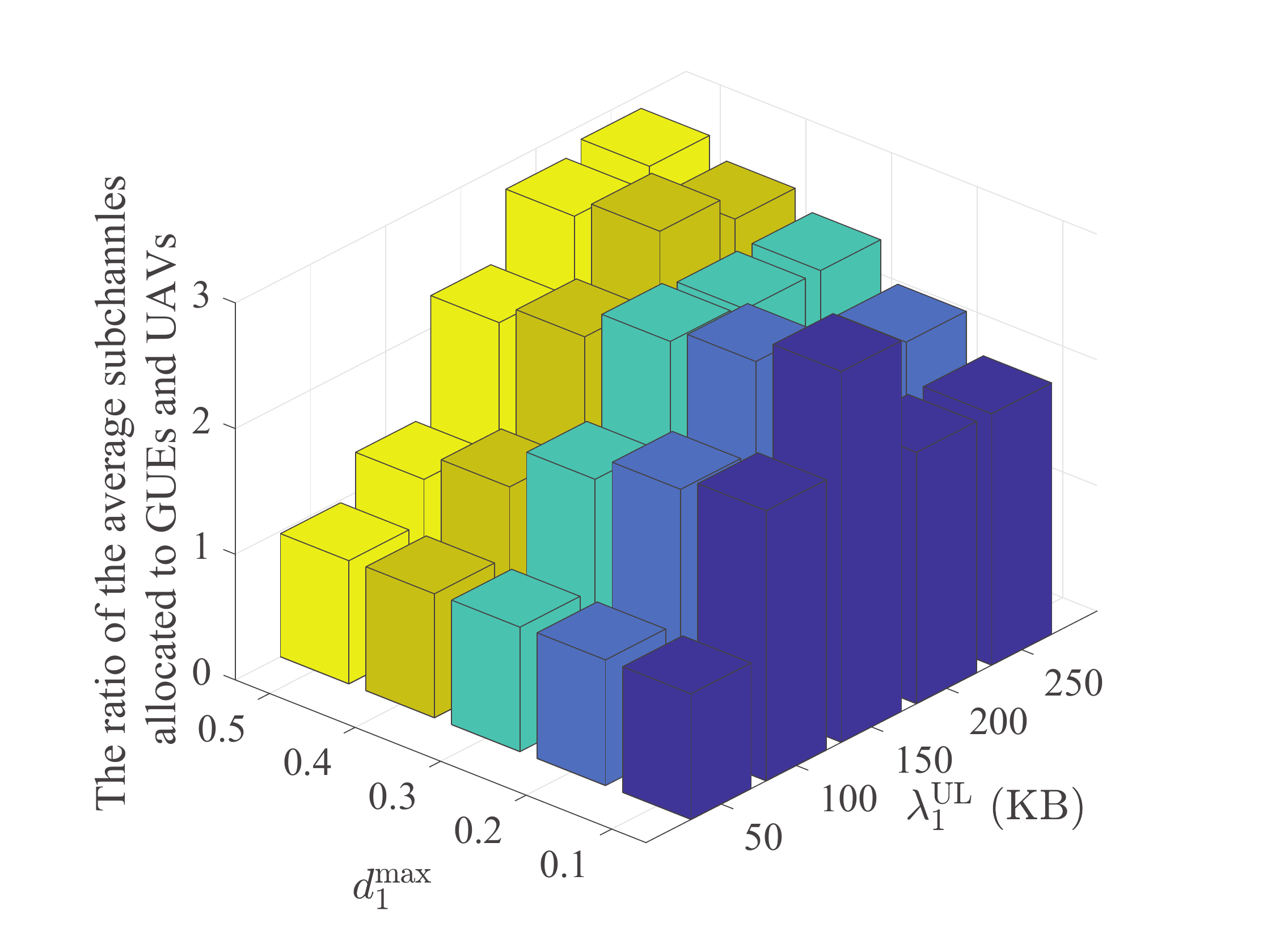}
\caption{ The ratio of average subchannels allocated to GUEs and UAVs ($\lambda_2^{\mathrm{UL}}=50$~KB, $d_2^{\max}=0.1$). }
\label{Fig. 7}
\end{figure}

\begin{figure}[htb]
\centering
\includegraphics[width=3.9 in]{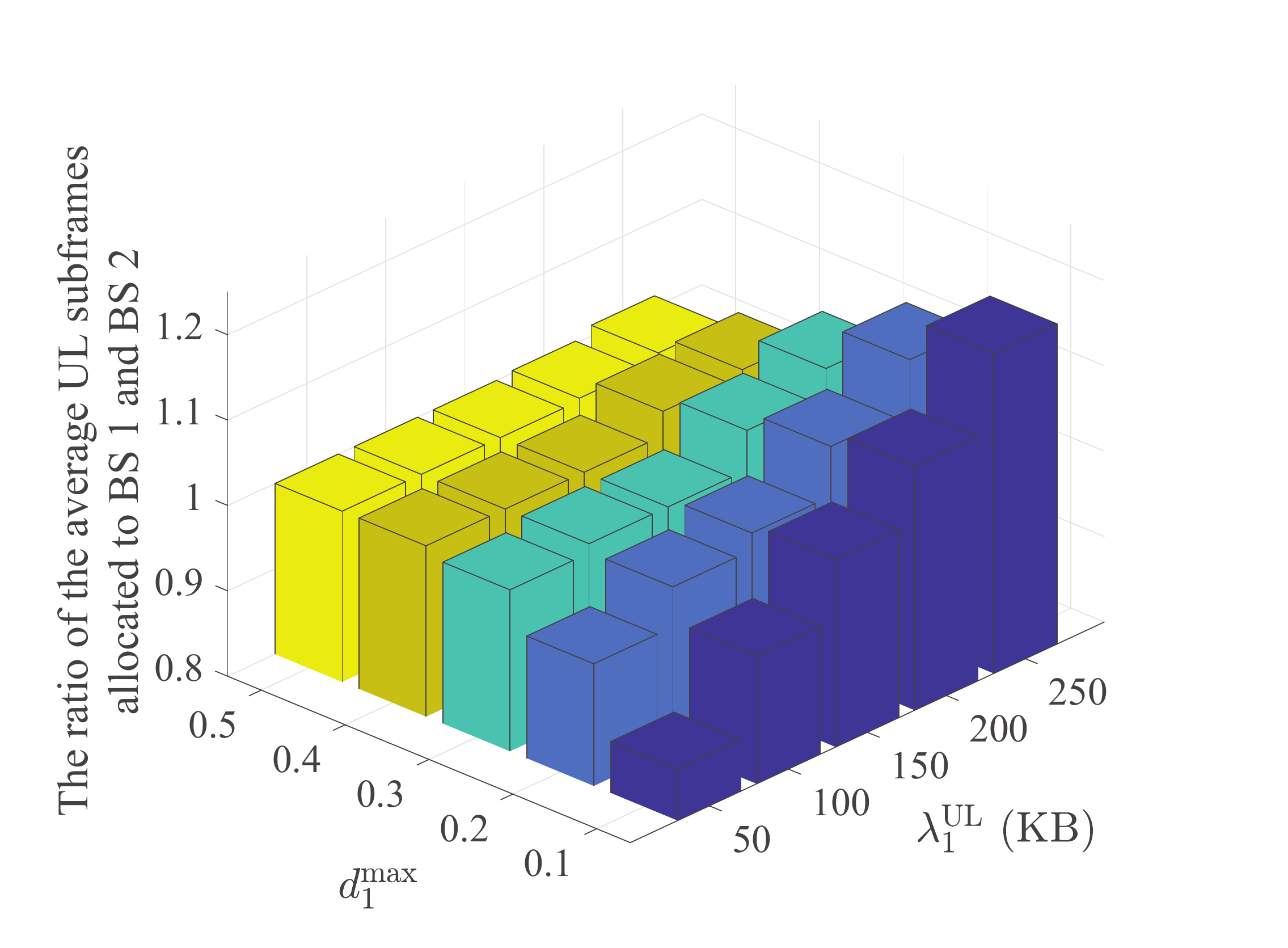}
\caption{ The ratio of average UL subframes allocated to BS 1 and BS 2 ($\lambda_1^{\mathrm{DL}}=200$ KB, $\lambda_2^{\mathrm{UL}}=50$ KB, $\lambda_2^{\mathrm{DL}}=80$ KB, $d_2^{\max}=0.1$). }
\label{Fig. 9}
\end{figure}

In Fig. \ref{Fig. 7} and Fig. \ref{Fig. 9}, we discuss the optimal policy for subchannel and subframe allocations. Without loss of generality, we consider a network composed of two BSs as a special case, where each BS allocates 5 subframes and 4 subchannels between two UEs.
In Fig. \ref{Fig. 7}, we study the effect of maximum tolerable dropping ratio and average packet arrival rate on the UL subchannel allocation policy, where the results can be extended to DL subchannel allocation. Next, we increase $d_1^{\max}$ and $\lambda_1^{\mathrm{UL}}$ to see the impact on the subchannel allocation ratio. On the one hand, when the average packet arrival rate $\lambda_1^{\mathrm{UL}}$ is relatively small (e.g., $\lambda_1^{\mathrm{UL}}\leq150$ KB), the number of subchannels allocated to GUE increases with $\lambda_1^{\mathrm{UL}}$. In this case, the QoS constraint of GUE is easily satisfied and thus the BS allocates more bandwidth resources to GUE to increase its rate. On the other hand, when $d_1^{\max}$ is relatively tight and $\lambda_1^{\mathrm{UL}}$ is relatively large (e.g., $d_1^{\max}=\{0.1,0.2\}$, $\lambda_1^{\mathrm{UL}}>150$ KB; $d_1^{\max}=0.4$, $\lambda_1^{\mathrm{UL}}\geq200$ KB), the number of subchannels allocated to GUE decreases with $\lambda_1^{\mathrm{UL}}$. In this case, it is difficult to satisfy the QoS constraint of GUE with heavy traffic load, thus the BS allocates more subchannels to UAV that has lighter data traffic. From the above discussions, we can see that the subchannel allocation needs to balance throughput and QoS constraints. When the bandwidth resources are sufficient, the BS prefers to allocate more sunchannels to the UEs with heavier data traffic loads for throughput enhancement. Otherwise,  it allocates fewer subchannels to those UEs whose QoS constraints are difficult to satisfy.

In Fig. \ref{Fig. 9}, we further  study the impact of the maximum tolerable dropping ratio and average packet arrival rate on the subframe allocation policy. To illustrate the asymmetric data traffic across the cells, we consider BS $1$ serves two GUEs and BS $2$ serves two UAVs, respectively. When the QoS constraint of $d_1^{\max}$ is relatively tight (e.g., $d_1^{\max}=\{0.1,0.2\}$), the number of UL subframes for BS 1  rapidly increases with the average UL packet size $\lambda_1^{\mathrm{UL}}$. For a large value of $\lambda_1^{\mathrm{UL}}$, we can see that the subframe allocation is unbalanced between the two BSs in order to meet the heavier UL data traffic demands for GUEs. Moreover, when $d_1^{\max}$ is relatively large (e.g., $d_1^{\max}=\{0.4,0.5\}$), two BSs have similar subframe configurations, which is to reduce inter-cell interference by controlling the number of unaligned subframes. From the above discussions, we can see that the subframe configuration needs to balance local traffic adaptation and inter-cell interference control. When the resources are sufficient, the BSs prefer to reduce the number of unaligned subframes for inter-cell interference control. In a resource-limited regime, each BS gives more priority to satisfying the local QoS constraints rather than interference avoidance.



\section{Conclusion}
In this paper, we proposed a user-centric D-TFDD scheme that fully utilizes both the time-domain and frequency-domain resources to meet the heterogeneous UEs' dynamic traffic demands while alleviating  inter-cell interference. Due to the limited observation space of the BSs, we formulated the D-TFDD control problem as a Dec-POMDP that maximizes the long-term expected sum rate of the network subject to the UEs' packet dropping ratio constraints. We proposed a federated reinforcement learning algorithm to solve this problem, where the BSs decide their local time-frequency configurations based on WDDPG algorithm and jointly update the global policy by exchanging the critic network parameters through  FL architecture.
Simulation results show that the proposed learning-based D-TFDD scheme is superior to other benchmark TDD schemes, and the proposed FWDDPG algorithm outperforms IDDPG and MADDPG algorithms by choosing the proper Wolpertinger coefficient. Our simulation results also reveal that, when the time-frequency resources are sufficient, the BS allocates more subchannels to the UEs with heavier traffic demands to improve the local data rate and adopts similar subframe configurations across the cells to mitigate inter-cell interference. In addition, in the resource-limited regime, the BS gives more priority to meeting local QoS constraints than to avoiding inter-cell interference.

\bibliographystyle{IEEEtran}
\bibliography{draftmulxin}
\begin{IEEEbiography}
[{\includegraphics[width=1.1in,height=1.3in,clip,keepaspectratio]{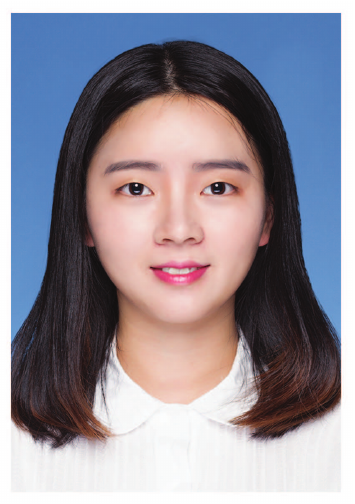}}]
{Ziyan Yin}(Student Member, IEEE) received the B.S. degree in the School of Electronic and Information Engineering from Suzhou University of Science and Technology, Suzhou, China, in 2017, and the M.Sc. degree from the School of Electronic and Optical Engineering, Nanjing University of Science and Technology, Nanjing, China, in 2018, where she is currently pursuing the Ph.D. degree. Her research interests include reinforcement learning, game theory, UAV communication and anti-jamming.
\end{IEEEbiography}
\begin{IEEEbiography}
[{\includegraphics[width=1.1in,height=1.2in,clip,keepaspectratio]{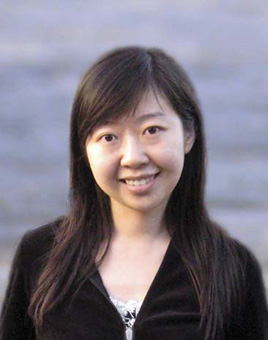}}]
{Zhe Wang}(Member, IEEE) received the Ph.D. degree in electrical engineering from The University of New South Wales, Sydney, Australia, in 2014. From 2014 to 2020, she was a research fellow with The University of Melbourne, Australia, and Singapore University of Technology and Design, Singapore, respectively. She is currently a professor with the School of Computer Science and Engineering, Nanjing University of Science and Technology, Nanjing, China. Her research interests include applications of optimization, game theory, and machine learning to resource allocation in communications and networking.
\end{IEEEbiography}

\begin{IEEEbiography}
[{\includegraphics[width=1in,height=1.25in,clip,keepaspectratio]{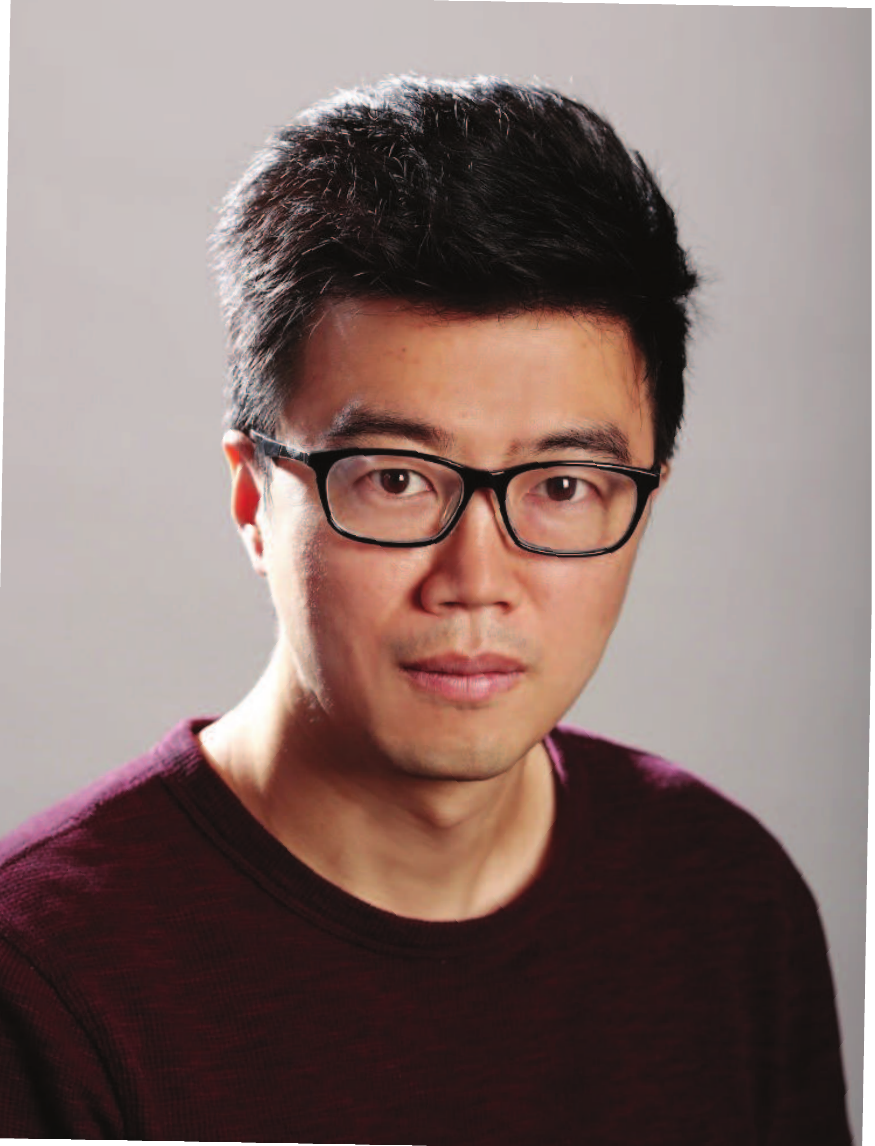}}]
{Jun Li}(Senior Member, IEEE) received Ph.D. degree in Electronic Engineering from Shanghai Jiao Tong University, Shanghai, P. R. China in 2009. From January 2009 to June 2009, he worked in the Department of Research and Innovation, Alcatel Lucent Shanghai Bell as a Research Scientist. From June 2009 to April 2012, he was a Postdoctoral Fellow at the School of Electrical Engineering and Telecommunications, the University of New South Wales, Australia. From April 2012 to June 2015, he was a Research Fellow at the School of Electrical Engineering, the University of Sydney, Australia. From June 2015 to now, he is a Professor at the School of Electronic and Optical Engineering, Nanjing University of Science and Technology, Nanjing, China. He was a visiting professor at Princeton University from 2018 to 2019. His research interests include network information theory, game theory, distributed intelligence, multiple agent reinforcement learning, and their applications in ultra-dense wireless networks, mobile edge computing, network privacy and security, and industrial Internet of things. He has co-authored more than 200 papers in IEEE journals and conferences, and holds 1 US patents and more than 10 Chinese patents in these areas. He is serving as an editor of IEEE Transactions on Wireless Communication and TPC member for several flagship IEEE conferences.
\end{IEEEbiography}
\begin{IEEEbiography}[{\includegraphics[width=1in,height=1.25in,clip,keepaspectratio]{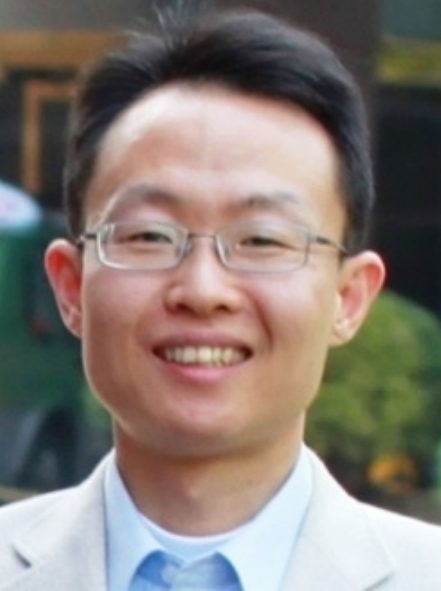}}]
{Ming Ding} (Senior Member, IEEE) received the B.S. and M.S. degrees (with first-class Hons.) in electronics engineering from Shanghai Jiao Tong University (SJTU), Shanghai, China, and the Doctor of Philosophy (Ph.D.) degree in signal and information processing from SJTU, in 2004, 2007, and 2011, respectively. From April 2007 to September 2014, he worked at Sharp Laboratories of China in Shanghai, China as a Researcher/Senior Researcher/Principal Researcher. Currently, he is a senior research scientist at Data61, CSIRO, in Sydney, NSW, Australia. His research interests include information technology, data privacy and security, machine learning and AI, etc. He has authored over 140 papers in IEEE journals and conferences, all in recognized venues, and around 20 3GPP standardization contributions, as well as a Springer book ``Multi-point Cooperative Communication Systems: Theory and Applications''. Also, he holds 21 US patents and co-invented another 100+ patents on 4G/5G technologies in CN, JP, KR, EU, etc. Currently, he is an editor of IEEE Transactions on Wireless Communications and IEEE Communications Surveys and Tutorials. Besides, he has served as Guest Editor/Co-Chair/Co-Tutor/TPC member for multiple IEEE top-tier journals/conferences and received several awards for his research work and professional services.
\end{IEEEbiography}

\begin{IEEEbiography}
[{\includegraphics[width=1in,height=1.25in,clip,keepaspectratio]{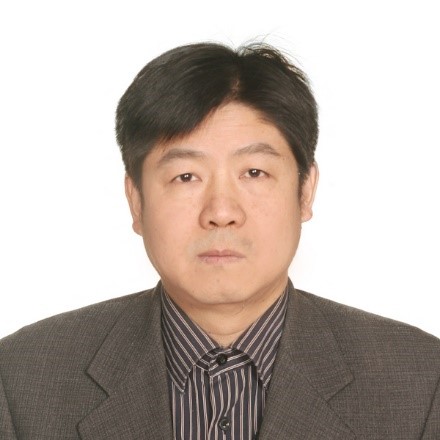}}]
{Wen Chen}(Senior Member, IEEE) is a tenured Professor with the Department of Electronic Engineering, Shanghai Jiao Tong University, China, where he is the director of Broadband Access Network Laboratory. He is a fellow of Chinese Institute of Electronics and the distinguished lecturers of IEEE Communications Society and IEEE Vehicular Technology Society. He is the Shanghai Chapter Chair of IEEE Vehicular Technology Society, Editors of IEEE Transactions on Wireless Communications, IEEE Transactions on Communications, IEEE Access and IEEE Open Journal of Vehicular Technology. His research interests include multiple access, wireless AI and meta-surface communications. He has published more than 120 papers in IEEE journals and more than 120 papers in IEEE Conferences, with citations more than 8000 in google scholar.
\end{IEEEbiography}
\begin{IEEEbiography}[{\includegraphics[width=1in,height=1.25in,clip,keepaspectratio]{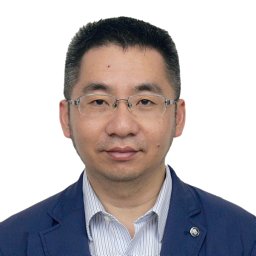}}]
{Shi Jin}(Senior Member, IEEE) received the B.S. degree in communications engineering from the Guilin University of Electronic Technology, Guilin, China, in 1996, the M.S. degree from the Nanjing University of Posts and Telecommunications, Nanjing, China, in 2003, and the Ph.D. degree in information and communications engineering from Southeast University, Nanjing, in 2007. From June 2007 to October 2009, he was a Research Fellow with the Adastral Park Research Campus, University College London, London, U.K. He is currently with the Faculty of the National Mobile Communications Research Laboratory, Southeast University. His research interests include space time wireless communications, random matrix theory, and information theory. He and his coauthors have been awarded the 2011 IEEE Communications Society Stephen O. Rice Prize Paper Award in the field of communication theory and the 2010 Young Author Best Paper Award by the IEEE Signal Processing Society. He serves as an Associate Editor for the IEEE Transactions on Communications, IEEE  Transactions on Wireless Communications, the IEEE Communications Letters, and IET Communications.
\end{IEEEbiography}

\end{document}